\documentclass{article}

\usepackage{arxiv}
\usepackage[utf8]{inputenc} % allow utf-8 input
\usepackage[T1]{fontenc}    % use 8-bit T1 fonts
\usepackage{hyperref}       % hyperlinks
\usepackage{url}            % simple URL typesetting
\usepackage[numbers]{natbib}
\usepackage{tabularx, etoolbox}
\usepackage{multirow}
\usepackage[capposition=top]{floatrow}
\usepackage{booktabs,siunitx}
\usepackage{enumitem}
\setlist[itemize]{leftmargin=.5cm}
\setlist[enumerate]{leftmargin=.5cm}
\setlist[description]{leftmargin=.5cm, labelindent=\parindent}
\usepackage[linesnumbered,ruled,vlined]{algorithm2e}
\usepackage{caption} 
\usepackage{amsmath}
\usepackage{graphicx}
\usepackage{xcolor}
\usepackage{listings}
\usepackage{booktabs}       % professional-quality tables
\usepackage[tableposition=top]{caption}
\usepackage[flushleft]{threeparttable}
\usepackage{amsfonts}       % blackboard math symbols
\usepackage{nicefrac}       % compact symbols for 1/2, etc.
\usepackage{microtype}      % microtypography
\usepackage{adjustbox}
\usepackage{textcomp}  % for compatibility of microtype and siunitx

\SetKwInput{KwInput}{Input}           
\SetKwInput{KwOutput}{Output}

\definecolor{halfgray}{gray}{0.55}
\definecolor{ipython_frame}{RGB}{207, 207, 207}
\definecolor{ipython_bg}{RGB}{247, 247, 247}
\definecolor{ipython_red}{RGB}{186, 33, 33}
\definecolor{ipython_green}{RGB}{0, 128, 0}
\definecolor{ipython_cyan}{RGB}{64, 128, 128}
\definecolor{ipython_purple}{RGB}{170, 34, 255}
\lstdefinestyle{ipython}{
    language=Python, 
    frame=lines,
    backgroundcolor=\color{white},
    commentstyle=\color{ipython_cyan},
    keywordstyle=\color{ipython_green},
    numberstyle=\ttfamily\tiny\color{halfgray},
    escapechar=\¢,escapebegin=\color{ipython_green},
    stringstyle=\color{ipython_red},
    basicstyle=\ttfamily\footnotesize,
    breakatwhitespace=false,
    breaklines=true,
    captionpos=t,
    keepspaces=true,
    numbers=left,
    numbersep=5pt,
    showspaces=false,
    showstringspaces=false,
    showtabs=false,
    tabsize=2,
    morekeywords={access,and,break,class,continue,def,del,elif,else,except,exec,finally,for,from,global,if,import,in,is,lambda,not,or,pass,print,raise,return,try,while},%
    morekeywords=[2]{abs,all,any,basestring,bin,bool,bytearray,callable,chr,classmethod,cmp,compile,complex,delattr,dict,dir,divmod,enumerate,eval,execfile,file,filter,float,format,frozenset,getattr,globals,hasattr,hash,help,hex,id,input,int,isinstance,issubclass,iter,len,list,locals,long,map,max,memoryview,min,next,object,oct,open,ord,pow,property,range,raw_input,reduce,reload,repr,reversed,round,set,setattr,slice,sorted,staticmethod,str,sum,super,tuple,type,unichr,unicode,vars,xrange,zip,apply,buffer,coerce,intern},%
    sensitive=true,%
    morecomment=[l]\#,%
    morestring=[b]',%
    morestring=[b]",%
    morestring=[s]{'''}{'''},% used for documentation text (mulitiline strings)
    morestring=[s]{"""}{"""},% added by Philipp Matthias Hahn
    morestring=[s]{r'}{'},% `raw' strings
    morestring=[s]{r"}{"},%
    morestring=[s]{r'''}{'''},%
    morestring=[s]{r"""}{"""},%
    morestring=[s]{u'}{'},% unicode strings
    morestring=[s]{u"}{"},%
    morestring=[s]{u'''}{'''},%
    morestring=[s]{u"""}{"""},%
    literate=
    *{+}{{{\color{ipython_purple}+}}}1
    {-}{{{\color{ipython_purple}-}}}1
    {*}{{{\color{ipython_purple}$^\ast$}}}1
    {/}{{{\color{ipython_purple}/}}}1
    {^}{{{\color{ipython_purple}\^{}}}}1
    {?}{{{\color{ipython_purple}?}}}1
    {!}{{{\color{ipython_purple}!}}}1
    {\%}{{{\color{ipython_purple}\%}}}1
    {<}{{{\color{ipython_purple}<}}}1
    {>}{{{\color{ipython_purple}>}}}1
    {|}{{{\color{ipython_purple}|}}}1
    {\&}{{{\color{ipython_purple}\&}}}1
    {~}{{{\color{ipython_purple}~}}}1
    {=}{{{\color{ipython_purple}=}}}1
    {==}{{{\color{ipython_purple}==}}}2
    {<=}{{{\color{ipython_purple}<=}}}2
    {>=}{{{\color{ipython_purple}>=}}}2
    {+=}{{{+=}}}2
    {-=}{{{-=}}}2
    {*=}{{{$^\ast$=}}}2
    {/=}{{{/=}}}2,
}
\robustify\bfseries
\newcommand\litem[1]{\item{\bfseries #1}} % new item enumeration

\newcommand{\InsertPercentageDifferenceTableNotes}[1]{\emph{Notes:} Rows show forecasters described in table \ref{tab:m4-models}. Columns show M4 data sets grouped by sampling frequency. Values show the percentage difference between replicated and published #1 values relative to the published values. Negative values indicate that replicated results are lower/better than published ones.
}

\newcommand{\InsertLRTableNotes}{We exclude LR model results when generated forecasts were instable/exploding, likely due the little available data and linear extrapolation.

}

\newcommand{\InsertDifferenceTableNotes}[1]{\emph{Notes:} Rows show forecasters described in table \ref{tab:m4-models}. Columns show M4 data sets grouped by sampling frequency. Values show the difference between replicated and published mean #1 values, together with the standard error of the difference in means between paired samples. Values in bold indicate that the difference is statistically significant at the $95\%$ level based on a two-sided paired t-test. Negative values indicate that replicated results are lower than published ones. 
}
\renewcommand{\keywords}[1]{{\small	\emph{Keywords:} #1}}

\title{
Forecasting with sktime: \\
Designing sktime's New Forecasting API and Applying It to Replicate and Extend the M4 Study
}

\author{
    Markus Löning\thanks{Corresponding author: \texttt{markus.loning@gmail.com}} \\
    University College London \\
    \And
    Franz J.~Kir\'{a}ly \\
    University College London \\
}

\begin{document}
\maketitle

%%%%%%%%%%%%%%%%%%%%%%%%%%%%%%%%%%%%%%%%%%%%%%%%%%%%%%%%%%%%%%%%%%%%%%%%%%%%%%%%%%%%
%%%%%%%%%%%%%%%%%%%%%%%%%%%%%%%%%%%%%%%%%%%%%%%%%%%%%%%%%%%%%%%%%%%%%%%%%%%%%%%%%%%%
\begin{abstract}
We present a new open-source framework for forecasting in Python. Our framework forms part of sktime, a more general machine learning toolbox for time series with scikit-learn compatible interfaces for different learning tasks. Our new framework provides dedicated forecasting algorithms and tools to build, tune and evaluate composite models. We use sktime to both replicate and extend key results from the M4 forecasting study. In particular, we further investigate the potential of simple off-the-shelf machine learning approaches for univariate forecasting. Our main results are that simple hybrid approaches can boost the performance of statistical models, and that simple pure approaches can achieve competitive performance on the hourly data set, outperforming the statistical algorithms and coming close to the M4 winner.
\end{abstract}

% keywords can be removed

%%%%%%%%%%%%%%%%%%%%%%%%%%%%%%%%%%%%%%%%%%%%%%%%%%%%%%%%%%%%%%%%%%%%%%%%%%%%%%%%%%%%
%%%%%%%%%%%%%%%%%%%%%%%%%%%%%%%%%%%%%%%%%%%%%%%%%%%%%%%%%%%%%%%%%%%%%%%%%%%%%%%%%%%%
\section{Introduction}
Time series forecasting is ubiquitous in real-world applications. Examples include forecasting of demand to fill up inventories, economic growth forecasts to inform policies, and predicting stock prices to guide financial decisions. Forecasting is also a fruitful area for machine learning research, and pure and hybrid machine learning approaches have recently achieved state-of-the-art performance \cite{Smyl2020, Oreshkin2019}.

In practice, forecasting involves a number of steps: we first need to specify, fit and select an appropriate model, and then evaluate and deploy it. There are various open-source toolboxes that help us implement these steps. However, most existing toolboxes are limited in important respects. Some support only specific model families (e.g.\ ARIMA or neural networks). Others provide more generic frameworks for forecasting, but no interfaces to existing machine learning toolboxes like scikit-learn \cite{Buitinck2013}. Still others offer functionality for only some steps of the steps (e.g.\ feature extraction). But despite the success of machine learning in forecasting, to our knowledge, there is no open-source toolbox that allows to interface existing machine learning toolboxes and to build, tune and evaluate composite machine learning models for forecasting. 

To close this gap, we present sktime's new forecasting framework in Python. We provide a composable and understandable forecasting interface and all the necessary functionality to build, tune and evaluate forecasting models. Our framework is embedded in sktime \cite{Loning2019}, a machine learning toolbox for time series that extends scikit-learn to different learning tasks that arise in a temporal data context, including forecasting, but also time series classification and regression among others. 

In this paper, we first motivate and describe the design of our forecasting framework. We then use it to replicate key results from the M4 forecasting study. In addition, we extend the M4 study by evaluating univariate machine learning models using sktime's off-the-shelf functionality for reduction, boosting, pipelining and tuning.

In our replication, we find no differences for the naïve models, small differences for statistical algorithms, and large improvements for machine learning models. In our extension, we find that simple hybrid machine learning models can boost the performance of statistical models, and that simple pure machine learning models can achieve competitive performance on the hourly data set, outperforming statistical models and coming close to the best M4 models. 

Finally, with sktime, we hope to further streamline open-source capabilities for machine learning with time series in Python, making algorithmic performance comparisons more transparent and reproducible.

%%%%%%%%%%%%%%%%%%%%%%%%%%%%%%%%%%%%%%%%%%%%%%%%%%%%%%%%%%%%%%%%%%%%%%%%%%%%%%%%%%%%
%%%%%%%%%%%%%%%%%%%%%%%%%%%%%%%%%%%%%%%%%%%%%%%%%%%%%%%%%%%%%%%%%%%%%%%%%%%%%%%%%%%%
\subsection*{Summary of contributions}
\begin{description}[
    style=unboxed,
    %leftmargin=0.5cm
]
    \item[sktime's forecasting framework.] To the best of our knowledge, we are the first to present an open-source machine learning toolbox for forecasting that allows to easily build, tune and evaluate composite machine learning models and is compatible with one of the major machine learning toolboxes (scikit-learn).
    
    \item[Replication and extension of the M4 study.] To our knowledge, we are the first to replicate the M4 study \cite{M4Team2018a, Makridakis2018a} and check the validity of the published results. In addition, we extend the M4 study by evaluating new machine learning approaches, including reduction, boosting, pipelining and tuning. %We also add tests to check whether found performance differences between algorithms are statistically significant. 
\end{description}

The remainder of the paper is organised as follows:
\begin{itemize}
    \item Section \ref{sec:problem} states the problems we are trying to solve with sktime's new forecasting framework.
    \item Section \ref{sec:api} motivates and describes the framework. 
    \item Section \ref{sec:literature} reviews related software and literature.
    \item Section \ref{sec:experiments} presents the results from replicating and extending the M4 study.
    \item Section \ref{sec:conclusion} concludes by suggesting future directions of research and development.
\end{itemize}  

%%%%%%%%%%%%%%%%%%%%%%%%%%%%%%%%%%%%%%%%%%%%%%%%%%%%%%%%%%%%%%%%%%%%%%%%%%%%%%%%%%%%
%%%%%%%%%%%%%%%%%%%%%%%%%%%%%%%%%%%%%%%%%%%%%%%%%%%%%%%%%%%%%%%%%%%%%%%%%%%%%%%%%%%%
\section{Problem statement}
\label{sec:problem}
We consider two problems: the \emph{practitioner's problem} of making accurate forecasts, and the \emph{developer's problem} of designing a good application programming interface (API) for solving the practitioner's problem.

\subsection{Forecasting}
\label{subsec:problem_forecasting}
For the practioner's problem, we consider the classical univariate forecasting problem with discrete time points. The task is to use the observations $\textbf{y} = (y(t_1)\dots y(t_T))$ of a single time series observed up to time point $t_T$ to find a forecaster $\hat{f}$ which can make accurate temporal forward predictions $\hat{y}(h_j) = \hat{f}(h_j)$ for the given time points $h_1\dots h_H$ of the forecasting horizon.\footnote{For an overview of the classical forecasting setting, see e.g.\ \cite{Box2015, Brockwell2016, Hyndman2018a, DeGooijer2006}} 
To evaluate the forecasting accuracy, we use performance metrics. Two common metrics are MASE (mean absolute scaled error) and sMAPE (symmetric mean absolute percentage error), as described in section \ref{sec:evaluation}. 

Note that we here assume equidistant time points, but our forecasting framework is flexible enough to support unequally-spaced series. It is also worth emphasising that we focus on univariate forecasting where only a single series is required for training. By contrast, many of the machine learning models submitted to the M4 study need multiple series for training. 

\subsection{API design}
The developer's problem is to find a good API to help solve the practitioner's forecasting problem, subject to a few extra requirements. Forecasting, like any other machine learning task, involves a number of mathematical concepts and operations. API design is about mapping these concepts and operations onto classes and methods in a programming language (here Python). 

Our extra requirements are that the API should be compatible with scikit-learn, so that we can re-use much of their functionality. Core functionality should also have a common interface, ensuring that the interface is modular and composable. This allows us to develop modular tools for building composite models that work with any forecaster or regressor. 

Assessing the goodness of an API is less straightforward than evaluating forecasting accuracy. Throughout the paper, we will make qualitative arguments to support our design choices drawing on the similarity to well established APIs, notably scikit-learn \cite{Buitinck2013}, and adherence to common design patterns and principles for object-oriented software development \cite{Erich2002}.

%%%%%%%%%%%%%%%%%%%%%%%%%%%%%%%%%%%%%%%%%%%%%%%%%%%%%%%%%%%%%%%%%%%%%%%%%%%%%%%%%%%%
%%%%%%%%%%%%%%%%%%%%%%%%%%%%%%%%%%%%%%%%%%%%%%%%%%%%%%%%%%%%%%%%%%%%%%%%%%%%%%%%%%%%
\section{Forecasting API}
\label{sec:api}

%%%%%%%%%%%%%%%%%%%%%%%%%%%%%%%%%%%%%%%%%%%%%%%%%%%%%%%%%%%%%%%%%%%%%%%%%%%%%%%%%%%%
%%%%%%%%%%%%%%%%%%%%%%%%%%%%%%%%%%%%%%%%%%%%%%%%%%%%%%%%%%%%%%%%%%%%%%%%%%%%%%%%%%%%
\subsection{Motivation}
\label{sec:motivation}
%Before we describe sktime's forecasting framework in detail, we want to briefly motivate why we develop sktime. 
Given the limited toolbox capabilities for time series analysis (see section \ref{sec:literature}), there are a number of reasons why we believe extending toolbox capabilities is important:

\begin{itemize}
    \litem{Rapid prototyping.} Toolboxes allow for rapid implementation and exploration of new models, allowing users and researchers to quickly and systematically evaluate and compare models. 
    \litem{Reproducibility.} Reproducibility is essential to scientific progress, and in particular to machine learning and forecasting research \cite{Makridakis2018c, Boylan2015, Hyndman2010, Arnold2019}. Toolboxes, like sktime, with a principled and modular interface, enable researchers to easily replicate results from available models and compare them against new models.
    \litem{Transparency.} By providing a consistent interface for algorithms and composition functionality, toolboxes make algorithms and workflows more readable and transparent, helping users and researchers to better understand how forecasts are generated.
\end{itemize}

In addition, there are a number of reasons why we develop our forecasting framework as part of sktime's unified API, as opposed to a separate forecasting toolbox: 
\begin{itemize}
    \litem{Reduce confusion.} When learning with time series, there are various related but distinct learning tasks (e.g.\ forecasting and time series classification). sktime's unified API is supported by a clear taxonomy of these tasks and corresponding types of algorithm that can solve them. As a result, model specification in sktime makes the task and algorithm type explicit. This avoids confusion about the task we are trying to solve and the types of algorithms we can use to solve it. Without such clear distinction, we risk conflating tasks and algorithm types. This may lead to inappropriate (or overly optimistic) algorithmic performance estimates and unfair performance comparisons. For example, it is often seen that performance estimates of the reduced regression setting are mistaken for performance estimates for the forecasting setting, which are in general not the same \cite{Bergmeir2018}.\footnote{The crucial difference is that in the regression setting, we usually assume samples to be independent whereas in the forecasting setting we cannot plausibly make such an assumption, as observations are dependent on past observations. We discuss reduction in more detail below and in section \ref{sec:reduction}.} Another example is given by the M4 study \cite{Makridakis2019} which includes both univariate and multivariate models, without distinguishing them explicitly. By univariate models we mean those models that use a single series for training (e.g.\ all of the statistical models in table \ref{tab:m4-models}), whereas multivariate use multiple series for training and hence can make use of consistent patters across series (e.g.\ the winner \cite{Smyl2020} and runner-up \cite{Montero-Manso2020a}). Comparing both univariate and multivariate models is problematic for two reasons: first, training models on multiple series introduces a dependency between the performance estimates on individual series, making them less reliable; second, it seems unfair to compare multivariate models with univariate ones, especially when this is not made explicit and if univariate models could easily make use of multivariate data too. We believe a unified API with a clear taxonomy of tasks and algorithm types will help reduce confusion.
    \litem{Reduction.} Many time series algorithms are highly composite and often involve reduction from a complex to a simpler learning tasks. Reduction relations exist between many time series related tasks, including forecasting and tabular (or cross-sectional) regression, but also time series regression, multivariate (or panel) forecasting, and time series annotation (e.g.\ anomaly detection) \cite{Loning2019}. Only a unified toolbox like sktime allows to fully exploit these relations. Reduction is discussed in more detail in section \ref{sec:reduction}.
    \litem{Re-usability.} Many time series learning tasks require common functionality (e.g.\ feature extraction, time series distances, or pre-processing routines). Providing them in a consistent and modular interface allows us to re-utilise them for different tasks.
\end{itemize}

%%%%%%%%%%%%%%%%%%%%%%%%%%%%%%%%%%%%%%%%%%%%%%%%%%%%%%%%%%%%%%%%%%%%%%%%%%%%%%%%%%%%
%%%%%%%%%%%%%%%%%%%%%%%%%%%%%%%%%%%%%%%%%%%%%%%%%%%%%%%%%%%%%%%%%%%%%%%%%%%%%%%%%%%%
\subsection{Basic forecaster interface}
\label{sec:forecaster}
We start discussing sktime's new forecasting framework by describing our basic interface for forecasting algorithms (or forecasters). We encapsulate forecasters in classes with a common interface, as is standard in existing toolboxes. The advantages of a common interface are clear: we can interchange forecasters at run-time and we can compose them, allowing us to write tools that work with any forecaster and to easily build composite models (e.g.\ ensembles or tuning routines). 
%We discuss composition in more detail in section \ref{sec:composition}. 
%
\begin{lstlisting}[
    caption=Base forecaster interface,
    label=py:base
]
forecaster = ExponentialSmoothing(trend="additive")
forecaster.fit(y_train)  # y_train is the training series
y_pred = forecaster.predict(fh=¢1¢)  # single-step ahead forecasting horizon (fh)
\end{lstlisting}
What is less obvious is what a common interface for forecasters should look like. We list the methods we consider essential in table \ref{tab:forecaster_api} and discuss them in more detail below. Example \ref{py:base} shows what our common interface looks like in practice.

\begin{table}[htbp]
    \centering \small
    \caption{Common forecaster interface}
    \label{tab:forecaster_api}
    \begin{tabularx}{\textwidth}{lXl}
 Functionality & Description & Method \\
 \toprule
 Specification & Building and initialising models, setting of hyper-parameters & \texttt{\_\_init\_\_} \\  
 Training & Fitting model parameters to training data & \texttt{fit}    \\
 Forecasting & Generating in-sample or out-of-sample predictions based on fitted parameters & \texttt{predict}\\
 Updating & Updating fitted parameters using new data & \texttt{update} \\
 Dynamic forecasting & Making and updating forecasts dynamically using temporal cross-validation & \texttt{update\_predict} \\
  Inspection & Retrieving hyper-parameters and fitted parameters & \texttt{get\_params}, \texttt{get\_fitted\_params}\\
%  Persistence & Saving fitted models & \texttt{save} \\
 \bottomrule
\end{tabularx}
\end{table}

% \begin{lstlisting}[
%     caption=Model evaluation, 
%     language=Python, 
%     label=py:evaluation
% ]
% f = ThetaForecaster() 
% f.fit(y_train)
% cv = SlidingWindowSplitter(fh=[¢1¢, ¢2¢, ¢3¢])  # multi-step ahead forecasts
% y_pred = f.update_predict(y_test, cv=cv)  # sliding-window cross-validation 
% \end{lstlisting}

\begin{itemize}
    \litem{Specification.} Like scikit-learn, but unlike statsmodels \cite{Perktold2010}, we separate model specification from the training data, following the general design principles of modularisation and decoupling. 
    % general principle of encapsulation, modularisation/decoupling, cohesion: class should focus on few things, logic emerges in interaction between classes with well-defined purposes; relation to mental model, anchors design patterns to mental/mathematical model, in relation to a certain kind of data, statistical generative model or learning task, but not actual data 
    \litem{Training.} Once specified, the model can take in training data for parameter fitting. Models that fit separate parameters for each step of the forecasting horizon will also require the forecasting horizon during training. 
    \litem{Forecasting horizon.} The forecasting horizon specifies the time points we want to predict. It could be specified in a number of ways. In sktime, we specify it as the steps ahead relative to the end of the training series. A relative horizon, as opposed to an absolute one like in statsmodels, has the advantage that it allows to update forecasts when time moves on without having to update the forecasting horizon. Specifying the forecasting horizon as an interval of time points as in statsmodels, or simply the number of steps ahead as in pmdarima, is not enough. Forecasters may fit separate parameters for each step and hence need to know the exact steps to avoid needless computations. A  consequence of our choice is that in-sample forecasts are specified as negative steps, going backwards from the end of the training series. Another consequence is that forecasters need to keep track of the last point of the training series, what we call the cutoff point. 
    \litem{Forecasting.} Once fitted, the forecaster can generate forecasts. We expose a single method for in-sample and out-of-sample forecasts, even though generating them may involve different routines. The advantage of a single method is that composition forecasters do not have to distinguish between different method calls of its component forecasters, and instead can delegate that decision to the component forecasters. We discuss detrending as an example of this case in section \ref{sec:detrend}.
    % As a consequence, the distinction between in-sample and out-of-sample forecasts now has to be encoded in the forecasting horizon itself, rather than the method. 
    \litem{Updating.} In addition to fitting, we introduce a method for updating forecasters with new data. This allows to keep track of the cutoff point as time moves on, but also to update fitted parameters without having to re-fit the whole model. 
    \litem{Dynamic forecasting.} We also introduce a method for making and updating forecasts more dynamically. This is useful for temporal cross-validation, where we generate and evaluate multiple forecasts based on different windows of the data. The method takes in test data and an iterator that encodes the temporal cross-validation scheme. 
    \litem{Inspection.} In addition to the common hyper-parameter interface from scikit-learn, we also propose a new uniform interface for fitted parameters. This enables us to have composite models which make use of fitted parameters of component models. We discuss feature extraction as a typical example in section \ref{sec:extract}.
\end{itemize}

% TODO Omar: My only comment would be that maybe it's worth adding a couple of paragraphs to section 3.2 about the prediction intervals and plotting APIs for forecasters? The former can be easy to omit or get wrong (I'm not sure the difference between prediction intervals and forecasting intervals is well-known outside of academic settings) but is important for model and forecasting introspection, and the latter provides a fast and convenient way to assess model performance, without needing to pollute the namespace with lots of extra imports. It also brings the prediction intervals and forecast into one place, which some developers may overlook if they implemented their own methods.

%%%%%%%%%%%%%%%%%%%%%%%%%%%%%%%%%%%%%%%%%%%%%%%%%%%%%%%%%%%%%%%%%%%%%%%%%%%%%%%%%%%%
%%%%%%%%%%%%%%%%%%%%%%%%%%%%%%%%%%%%%%%%%%%%%%%%%%%%%%%%%%%%%%%%%%%%%%%%%%%%%%%%%%%%
\subsection{Composition} 
\label{sec:composition}
With the common forecaster interface in place, we propose a number of composition forecasters that enable us to build composite models based on one or more component forecasters (e.g.\ ensembles). Through these composition interfaces sktime enables to build a wide variety of models with a small amount of easy-to-read code. As is standard, composition forecasters (or meta-forecasters) are forecasters themselves, and hence share the basic interface. This allows us treat simple and composite forecasters uniformly. Our composition forecasters include adaptations of common tabular meta-estimators from scikit-learn to the forecasting setting, like pipelining, ensembling and tuning, but also novel meta-forecasters for reduction, detrending and feature extraction. 

%%%%%%%%%%%%%%%%%%%%%%%%%%%%%%%%%%%%%%%%%%%%%%%%%%%%%%%%%%%%%%%%%%%%%%%%%%%%%%%%%%%%
%%%%%%%%%%%%%%%%%%%%%%%%%%%%%%%%%%%%%%%%%%%%%%%%%%%%%%%%%%%%%%%%%%%%%%%%%%%%%%%%%%%%
\subsubsection{Reduction}
\label{sec:reduction}

As described in section \ref{sec:motivation}, one of the main reasons for developing a unified API is reduction, i.e.\ the insight that algorithms that can solve one task, can also be used to solve another task. Many machine learning approaches to forecasting work through reduction. 

For example, a common approach is to solve forecasting via regression. We typically do this as follows: we first split the training series into fixed-length windows and stack them on top of each other. This gives us a matrix of lagged values in a tabular format, and thus allows us to apply any tabular regression algorithm. 
%This approach is sometimes also called lagged variable regression, dynamic regression or auto-regression. 
To generate forecasts, there are multiple strategies, a common one is the recursive strategy. Here we use the last window as input to the fitted regressor to generate the first step ahead forecast. To make multi-step ahead forecasts, we can update the last window recursively with the previously forecasted values. Other strategies are the direct and hybrid strategies (for more details, see \cite{Bontempi2012}).

While reductions are not new, we are the first to propose encapsulating them as meta-estimators. Reductions have several key properties that make them well suited to be expressed as meta-estimators:
\begin{itemize}
    \litem{Modularity.} Reductions convert any algorithm for a particular task into an algorithm for a new task. Applying some reduction approach to $n$ base algorithms gives $n$ new algorithms for the new task. Any progress on the base algorithm immediately transfers to the new task, saving both research and software development effort \cite{Beygelzimer2005, Beygelzimer2008}. 
    \litem{Tunability.} Most reductions require modelling choices that we may want to optimise. For example, we may want to tune the window length or select among different strategies for generating forecasts \cite{Taieb2014, Bontempi2012}. By expressing reductions as meta-estimators, we expose these choices via the common interface as tunable hyper-parameters. 
    \litem{Composability.} Reductions are composable. They can be composed to solve more complicated problems \cite{Beygelzimer2005, Beygelzimer2008}. For example, we can first reduce forecasting to time series regression which in turn can be reduced to tabular regression via feature extraction.
    \litem{Adaptor.} Reductions adapt the interface of the base algorithm to the interface required for solving the new task, allowing us to use the common tuning and model evaluation tools appropriate for the new task. 
\end{itemize}
Due to the current lack of a unified toolbox, reductions are often hand-crafted, the M4 study being a case in point. The consequence is that they are neither adaptors, nor modular, tunable or composable. Example \ref{py:reduce} shows what reduction to tabular regression looks like in sktime, and we make heavy use of it in section \ref{sec:experiments} to replicate and extend the M4 study. We also provide a meta-forecaster for reduction to time series regression, so that any of sktime's time series regressors can be used to solve a forecasting task.
\begin{lstlisting}[
caption=Solving forecasting via reduction to tabular regression, 
label=py:reduce
]
regressor = RandomForestRegressor()  # from scikit-learn
forecaster = ReducedRegressionForecaster(regressor, window_length=¢10¢, strategy="recursive")
forecaster.fit(y_train)
y_pred = forecaster.predict(fh=¢1¢)
\end{lstlisting}

%%%%%%%%%%%%%%%%%%%%%%%%%%%%%%%%%%%%%%%%%%%%%%%%%%%%%%%%%%%%%%%%%%%%%%%%%%%%%%%%%%%%
%%%%%%%%%%%%%%%%%%%%%%%%%%%%%%%%%%%%%%%%%%%%%%%%%%%%%%%%%%%%%%%%%%%%%%%%%%%%%%%%%%%%
\subsubsection{Detrending}
\label{sec:detrend}
sktime provides a number of transformers which allow to apply data transformations. Similar to scikit-learn, they share a common interface for fitting, transforming and, if available, the inverse transformation. In contrast to scikit-learn's transformers, the transformers presented here operate on a single series. But sktime provides modular functionality to apply the single-series transformers on data frames with multiple series, so that they are re-usable for different learning tasks.

In particular, we introduce a new modular detrending transformer, a composite transformer which works with any forecaster. It works by first fitting the forecaster to the input data. To transform data, it uses the fitted forecaster to generate forecasts for the time points of the passed data and returns the residuals of the forecasts. Depending on the passed data, this will require to generate in-sample or out-of-sample forecasts. Example \ref{py:detrend} shows how we can use the detrending transformer to remove a linear trend from the time series. 
\begin{lstlisting}[caption=Detrending, label=py:detrend]
forecaster = PolynomialTrendForecaster(degree=¢1¢)
transformer = Detrender(forecaster)  # linear detrending
transformer.fit(y_train)
yt = transformer.transform(y_train)  # returns in-sample residuals
\end{lstlisting}

The detrender also works in a pipeline as a form of boosting, by first detrending a time series and then fitting another forecaster on the residuals \cite{Taieb2014b}. We investigate the potential of boosting a statistical method with machine learning algorithms in section \ref{sec:experiments_extend}.

%%%%%%%%%%%%%%%%%%%%%%%%%%%%%%%%%%%%%%%%%%%%%%%%%%%%%%%%%%%%%%%%%%%%%%%%%%%%%%%%%%%%
%%%%%%%%%%%%%%%%%%%%%%%%%%%%%%%%%%%%%%%%%%%%%%%%%%%%%%%%%%%%%%%%%%%%%%%%%%%%%%%%%%%%
\subsubsection{Pipelining}
Following scikit-learn, we provide a composition forecaster for chaining one or more transformers with a final forecaster. When fitting the pipeline, the data is first transformed before being passed to the forecaster. To make forecasts, the forecaster first generates forecasts which are then inverse-transformed before being returned. Since the transformers work on the target series to be forecasted, we follow scikit-learn in calling this meta-estimator  \texttt{TransformedTargetForecaster}. 

Example \ref{py:pipe} shows how the Naïve2 strategy from the M4 study, described in table \ref{tab:m4-models}, can be expressed as a pipeline of a deseasonalisation step and a naïve forecaster. But note that our implementation allows to chain multiple transformations.
\begin{lstlisting}[caption=Pipeline, label=py:pipe]
forecaster = TransformedTargetForecaster([
    ("deseasonalise", Deseasonaliser(sp=¢12¢)),  # monthly seasonal periodicity
    ("forecast", NaiveForecaster(strategy="last"))
])
forecaster.fit(y_train)
y_pred = forecaster.predict(fh=¢1¢)
\end{lstlisting}

%%%%%%%%%%%%%%%%%%%%%%%%%%%%%%%%%%%%%%%%%%%%%%%%%%%%%%%%%%%%%%%%%%%%%%%%%%%%%%%%%%%%
%%%%%%%%%%%%%%%%%%%%%%%%%%%%%%%%%%%%%%%%%%%%%%%%%%%%%%%%%%%%%%%%%%%%%%%%%%%%%%%%%%%%
\subsubsection{Ensembling}
Following scikit-learn, we provide a simple meta-forecaster for ensembling multiple base forecasters. The ensemble forecaster fits each component forecaster separately and combines forecasts using a simple arithmetic mean. Given sktime's modular structure, it is straightforward to add other approaches to combine forecasts like weighted averages or stacking \cite{Chan2018, Smith2009, Timmermann2006, Clemen1989, Bates1969}.  

%%%%%%%%%%%%%%%%%%%%%%%%%%%%%%%%%%%%%%%%%%%%%%%%%%%%%%%%%%%%%%%%%%%%%%%%%%%%%%%%%%%%
%%%%%%%%%%%%%%%%%%%%%%%%%%%%%%%%%%%%%%%%%%%%%%%%%%%%%%%%%%%%%%%%%%%%%%%%%%%%%%%%%%%%
\subsubsection{Feature extraction}
\label{sec:extract}
Forecasting algorithms can not only be used to solve forecasting tasks, but can also help solve other related learning tasks. A common approach is to use forecasting algorithms as a feature extraction method for solving tasks such as time series regression, classification or clustering. %These tasks do not require temporal forward predictions of the same series, but instead use time series as features to predict some other target variable. 
This works by first fitting a forecaster to the available time series, then retrieving their fitted parameters, and finally using them as features for some tabular estimator. There are both bespoke models which make use of this approach (see e.g.\ the random interval spectral ensemble \cite{Bagnall2017} for time series classification, which makes use of auto-regressive coefficients) and toolkits like tsfresh \cite{Christ2016, Christ2018}) which allow to extract numerous features from time series, including fitted parameters from certain forecasting algorithms. 

To allow for more configurable feature extraction, we propose a feature extraction transformer, which is a meta-estimator that extracts the fitted parameters from a forecaster. To ensure full modularity of the transformer, we propose a new common inspection interface for retrieving fitted parameters in a uniform manner, as described in section \ref{sec:forecaster}. Example \ref{py:extract} shows how this transformer could be used in a pipeline for time series classification. 
\begin{lstlisting}[caption=Feature extraction for classification, label=py:extract]
forecaster = ARIMA()
classifier = Pipeline([
    ("extract", FittedParamExtractor(forecaster, params=["ar_params"])),
    ("classify", RandomForestClassifier())
])
classifier.fit(X_train, y_train)  # sktime's time series classification framework
y_pred = classifier.predict(X_test)
\end{lstlisting}

To our knowledge, we are the first to propose a common interface for fitted parameters, but we strongly encourage, and hope, that other toolboxes like scikit-learn will follow us. This would allow to extract features from composite models with scikit-learn components and open a number of other possibilities for model composition.

%%%%%%%%%%%%%%%%%%%%%%%%%%%%%%%%%%%%%%%%%%%%%%%%%%%%%%%%%%%%%%%%%%%%%%%%%%%%%%%%%%%%
%%%%%%%%%%%%%%%%%%%%%%%%%%%%%%%%%%%%%%%%%%%%%%%%%%%%%%%%%%%%%%%%%%%%%%%%%%%%%%%%%%%%
\subsection{Model selection}
Similar to scikit-learn, we have a tuning meta-forecaster. It performs grid-search cross-validation based on cross-validation iterator encoding the cross-validation scheme, the parameter grid to search over, and optionally the evaluation metric for comparing model performance. As in scikit-learn, tuning works through the common hyper-parameter interface which allows to repeatedly fit and evaluate the same forecaster with different hyper-parameters. 
\begin{lstlisting}[caption=Model selection, label=py:tune]
forecaster = ReducedRegressionForecaster(RandomForestRegressor(), window_length=¢3¢)
param_grid = {"window_length": [¢3¢, ¢5¢, ¢7¢]}
cv = SlidingWindowSplitter()  # cross-validation object
gscv = ForecastingGridSearchCV(forecaster, param_grid, cv)
gscv.fit(y_train)  # performs temporal grid-search CV 
y_pred = gscv.predict(fh=¢1¢)  # makes predictions based on best model found via CV
\end{lstlisting}

%%%%%%%%%%%%%%%%%%%%%%%%%%%%%%%%%%%%%%%%%%%%%%%%%%%%%%%%%%%%%%%%%%%%%%%%%%%%%%%%%%%%
%%%%%%%%%%%%%%%%%%%%%%%%%%%%%%%%%%%%%%%%%%%%%%%%%%%%%%%%%%%%%%%%%%%%%%%%%%%%%%%%%%%%
\subsection{Technical details}
\label{sec:sktime_technical}
sktime is available via PyPI and can be installed using Python's package manager \texttt{pip}. We distribute compiled files for Windows, MacOS and Linux for easy installation. The forecasting framework is available starting from version $0.4.0$.

sktime requires Python $3.6$ or later, and has a number of core dependencies, including: NumPy \cite{VanderWalt2011, Oliphant2015}, pandas \cite{McKinney2011} for data handling; SciPy \cite{Jones2001, Haenel2013}, scikit-learn \cite{Buitinck2013, Varoquaux2015, Pedregosa2001}, statsmodels \cite{Perktold2010} for statistical methods; and numba \cite{Lam2015}, Cython \cite{behnel2011cython} and joblib\footnote{\url{https://github.com/joblib/joblib}} for optimizations. For deep learning, sktime has a companion package, called sktime-dl\footnote{\url{https://github.com/sktime/sktime-dl}}, based on TensorFlow \cite{Abadi2016} and Keras \cite{Chollet2015}. 

We use continuous integration services for unit testing and code quality checks. We have extensive online documentation with interactive tutorials on Binder \cite{Bussonnier2018}, allowing users to try out sktime without having to install sktime. sktime is distributed under a permissive BSD-3-clause license and an active open-source community. We are looking for new contributors, and contributors can help improve and maintain existing functionality or lead the development of new frameworks. 

%%%%%%%%%%%%%%%%%%%%%%%%%%%%%%%%%%%%%%%%%%%%%%%%%%%%%%%%%%%%%%%%%%%%%%%%%%%%%%%%%%%%
%%%%%%%%%%%%%%%%%%%%%%%%%%%%%%%%%%%%%%%%%%%%%%%%%%%%%%%%%%%%%%%%%%%%%%%%%%%%%%%%%%%%
\section{Related work}
\label{sec:literature}

\subsection{Related software}
There are various well-developed toolboxes for the tabular (or cross-sectional) setting, which have established key design patterns for machine learning APIs: most notably, scikit-learn \cite{Varoquaux2015, Buitinck2013, Pedregosa2001} in Python, Weka \cite{Holmes1994, Hall2009} in Java, MLJ \cite{Blaom2020} in Julia, and mlr \cite{Bischl2016} or caret \cite{Kuhn2008, Kuhn2018} in R, all of which implement common interfaces for fitting, predicting and hyper-parameters, and support composite model building and tuning.

% TODO: update MLJ reference when paper is published

Beyond the cross-sectional setting, toolbox capabilities remain limited.\footnote{For a regularly updated and more extensive overview of Python libraries for time series analysis, see \url{https://github.com/alan-turing-institute/sktime/wiki/Related-software}.} There are a few toolboxes that extend tabular toolboxes and provide frameworks for time series learning tasks closely related to the cross-sectional setting, such as time series classification, regression and clustering. This includes pyts \cite{Faouzi2020}, seglearn \cite{Burns2018} %\texttt{seqlearn}\footnote{\url{https://github.com/larsmans/seqlearn}}
and tslearn\footnote{\url{https://github.com/rtavenar/tslearn}} in Python and tsml\footnote{\url{https://github.com/uea-machine-learning/tsml/}} \cite{Bagnall2017} in Java. However, none of them have a dedicated forecasting API. Other toolboxes extend tabular toolboxes by providing functionality to solve specific steps of a time series modelling workflow, most prominently, feature extraction toolboxes such as tsfresh \cite{Christ2018, Christ2016}, Featuretools \cite{Kanter2015} and hctsa \cite{Lubba2019, Fulcher2017, Fulcher2014}. In addition, there are a number of smaller toolkits for specific reduction approaches from tabular toolboxes to different time series learning tasks, such as time series regression and forecasting \cite{Hamilton2017}.

%TODO update tslearn reference once published

There are also a few toolboxes specifically for forecasting. However, most of them have important limitations. Arguably one of the most popular and comprehensive toolboxes for forecasting is the forecast library \cite{Hyndman2008, Hyndman2018a} in R. Together with its companion libraries, forecast provides extensive functionality for statistical and encapsulated machine learning algorithms, as well as for pre-processing, model selection and evaluation. Similarly, gluonts \cite{Alexandrov2019} in Python provides deep-learning models for probabilistic forecasting and interfaces other packages like forecast. But both are limited in their support for composite model building and do not integrate with available machine learning libraries like scikit-learn. Other forecasting toolboxes in Python are further limited to specific model families. statsmodels \cite{Perktold2010} provides extensive tools for time series analysis, including forecasting, but is limited to statistical models (e.g.\ ARIMA, exponential smoothing and state space models). pmdarima \cite{Smith2019} ports forecast's Auto-ARIMA algorithm \cite{Hyndman2008} into Python and provides additional tools for seasonality testing, pre-processing and pipelining, but is limited to the ARIMA family. Similarly, PyFlux \cite{Taylor2016a} is limited to generalised auto-regressive models (e.g.\ GARCH, GAS), and fbprophet \cite{Taylor2018} to general additive models. 

Finally, there are a number of repositories which collect and combine popular forecasting models via interfaces to existing libraries with tools to automate workflows, such as atspy \cite{Snow2020} and the Microsoft forecasting repository\footnote{\url{https://github.com/microsoft/forecasting}}, but none of them support composite model building.

%%%%%%%%%%%%%%%%%%%%%%%%%%%%%%%%%%%%%%%%%%%%%%%%%%%%%%%%%%%%%%%%%%%%%%%%%%%%%%%%%%%%
%%%%%%%%%%%%%%%%%%%%%%%%%%%%%%%%%%%%%%%%%%%%%%%%%%%%%%%%%%%%%%%%%%%%%%%%%%%%%%%%%%%%
\subsection{Related literature}
There is a long history of empirical comparison of forecasting algorithms. The M4 study \cite{Makridakis2018a, M4Team2018a} is the latest in an influential series of forecasting competitions organised by Spyros Makridakis since 1982 \cite{Makridakis1982}, with the fifth edition currently running on Kaggle.\footnote{\url{https://www.kaggle.com/c/m5-forecasting-accuracy/overview}} Previous competitions include one on energy demand \cite{Hong2016}, one on tourism data \cite{Athanasopoulos2011}, and the M3 competition \cite{Makridakis2000, Crone2011, Ahmed2010}. In addition, several articles have reviewed the competition results, including a special issue of the International Journal of Forecasting \cite{Makridakis2018, Fry2020, Gilliland2020, Bontempi2020, Taieb2014}. While machine learning approaches have received special attention in all of the previous competitions, they have also been reviewed in \cite{Zhang2012, Zhang1998, Hewamalage2019} with a focus on deep learning.

%%%%%%%%%%%%%%%%%%%%%%%%%%%%%%%%%%%%%%%%%%%%%%%%%%%%%%%%%%%%%%%%%%%%%%%%%%%%%%%%%%%%
%%%%%%%%%%%%%%%%%%%%%%%%%%%%%%%%%%%%%%%%%%%%%%%%%%%%%%%%%%%%%%%%%%%%%%%%%%%%%%%%%%%%
\section{Experiments: Replicating \& extending the M4 study}
\label{sec:experiments}
We use sktime's forecasting framework to replicate and extend the M4 study. This allows us to test our algorithm implementations and to showcase the usefulness of our framework. In addition, we can cross-check published results from the M4 study and further investigate the potential of machine learning models for forecasting.

%%%%%%%%%%%%%%%%%%%%%%%%%%%%%%%%%%%%%%%%%%%%%%%%%%%%%%%%%%%%%%%%%%%%%%%%%%%%%%%%%%%%
%%%%%%%%%%%%%%%%%%%%%%%%%%%%%%%%%%%%%%%%%%%%%%%%%%%%%%%%%%%%%%%%%%%%%%%%%%%%%%%%%%%%
\subsection{Data}
We use the 100k-series data set of the M4 study provided by \cite{Makridakis2018a, M4Team2018, M4Team2018a}. The data set consists of data frequently encountered in business, financial and economic forecasting. The series are grouped by sampling frequency into yearly, quarterly, monthly, weekly, daily and hourly data sets. Tables \ref{tab:m4-domain} and \ref{tab:m4-training} in the appendix present summary statistics, showing wide variability in time series characteristics and lengths of the available training series. 

%%%%%%%%%%%%%%%%%%%%%%%%%%%%%%%%%%%%%%%%%%%%%%%%%%%%%%%%%%%%%%%%%%%%%%%%%%%%%%%%%%%%
%%%%%%%%%%%%%%%%%%%%%%%%%%%%%%%%%%%%%%%%%%%%%%%%%%%%%%%%%%%%%%%%%%%%%%%%%%%%%%%%%%%%
\subsection{Model evaluation}
\label{sec:evaluation}
We only evaluate point forecasts in this paper. To evaluate the accuracy of point forecasts on a single series, we use sMAPE and MASE:
\begin{align*}
    \text{sMAPE} &= \frac{200}{H} \sum_{i=1}^{H} \frac{| y(h_{i}) - \hat{y}(h_{i}) | }{|y(h_{i})| + |\hat{y}(h_{i})|} &
    \text{MASE} &= \frac{1}{H} \sum_{i=1}^{H} \frac{| y(h_{i}) - \hat{y}(h_{i}) | }{\frac{1}{T + H - m} \sum_{j=m+1}^{T + H} |y(t_j) - y(t_{j-m})|}
\end{align*}
where the denominator of MASE is the naïve seasonal in-sample forecasts and $m$ the seasonal periodicity (or periods per year) of the data (e.g.\ 12 for monthly data). MASE and sMAPE are scale-independent metrics and hence appropriate for comparing forecasting algorithms across different series \cite{Hyndman2006}. 

In addition, we use OWA (overall weighted average), which is used in the M4 study to rank entries. OWA is an aggregate performance metric over multiple series:
\begin{align*}
    \text{OWA} &= \frac{1}{2} \left[ \frac{\frac{1}{N}\sum^{N}_{i}{\text{sMAPE}_i}}{\frac{1}{N}\sum^{N}_{i}\text{sMAPE}_{i, \text{Naïve2}}} + 
    \frac{\frac{1}{N}\sum^{N}_{i}\text{MASE}_i}{\frac{1}{N}\sum^{N}_{i}\text{MASE}_{i, \text{Naïve2}}} \right]
\end{align*}
where $N$ is the number of time series we aggregate over, the subscript $i$ denotes the index of an individual series, and $\text{sMAPE}_{i, \text{Naïve2}}$ and $\text{MASE}_{i, \text{Naïve2}}$ are the respective metrics for series $i$ and the Naïve2 forecaster described in table \ref{tab:m4-models}. 

In addition, we test if found performance differences are statistically significant given the variation in performance over individual series. First, we test whether the replicated results are significantly different from published results using paired t-tests. Note that standard errors are not computed based on replicates, i.e.\ multiple runs of same forecaster on the same data set, but based on a single run on multiple data sets. Given the nature of the M4 data set, we cannot plausibly assume that the series are independently and identically distributed. Consequently, the test results have to be interpreted with caution. 

Second, we extend the previous statistical analysis of the M4 results \cite{Makridakis2019, Koning2005} to check whether found performance differences between models are statistically significant. We use Friedman tests to check whether the found average sMAPE ranks are significantly different from the mean rank expected under the null-hypothesis at the $5$\% level. We then use post-hoc Nemenyi tests to find those pairs of models that are significantly different, where model pairs are defined by the originally published and replicated performance estimates. We summarise our findings visually in critical difference diagrams, as proposed by \cite{Demsar2006}. To validate our findings, we also use pairwise Wilcoxon signed rank tests together with Holm's correction procedure for multiple testing, as recommended by \cite{Garcia2008, Benavoli2016}.

%%%%%%%%%%%%%%%%%%%%%%%%%%%%%%%%%%%%%%%%%%%%%%%%%%%%%%%%%%%%%%%%%%%%%%%%%%%%%%%%%%%%
%%%%%%%%%%%%%%%%%%%%%%%%%%%%%%%%%%%%%%%%%%%%%%%%%%%%%%%%%%%%%%%%%%%%%%%%%%%%%%%%%%%%
\subsection{Technical implementation}
\label{sec:technical}
The code for replicating and extending the M4 study can be found on GitHub.\footnote{\url{https://github.com/mloning/sktime-m4}} We ran the experiments on machines with Linux CentOS $7.4$, $32$ CPUs and $189$ GB RAM. 

For all forecasters and composition tools, we use sktime. Forecasters are specified as composite models whenever possible, using the composition classes described in section \ref{sec:api}. For all regressors except XGB and RNN, we use scikit-learn \cite{Buitinck2013}. For XGB, we use xgboost \cite{Chen2016}. For RNN, we use skime-dl (see section \ref{sec:sktime_technical}).

%%%%%%%%%%%%%%%%%%%%%%%%%%%%%%%%%%%%%%%%%%%%%%%%%%%%%%%%%%%%%%%%%%%%%%%%%%%%%%%%%%%%
%%%%%%%%%%%%%%%%%%%%%%%%%%%%%%%%%%%%%%%%%%%%%%%%%%%%%%%%%%%%%%%%%%%%%%%%%%%%%%%%%%%%
\subsection{Replicating the M4 study}
\label{subsubsec:results_replicate}

\subsubsection{Models}
To replicate key results from the M4 study, we implement and re-evaluate all baseline forecasters of the M4 study in sktime, except the automatic exponential smoothing model (ETS). We also evaluate the improved Theta model by Legaki \& Koutsouri, the best statistical model in the M4 study. We give an overview of the replicated forecasters in table \ref{tab:m4-models} in the appendix. 

\subsubsection{Results}
For each model, we compare our findings against published results. We focus on average performance and computational run time. 

Our main results are presented in table \ref{tab:smape_perc_diff}, which shows the percentage differences between replicated and published sMAPE values for the data sets grouped by sampling frequency. Corresponding results for MASE and OWA are shown in the appendix in tables \ref{tab:mase_perc_diff} and \ref{tab:owa_perc_diff}. We also test whether the found differences are statistically significant using a paired t-test. Detailed results of the significance tests for sMAPE and MASE values are shown in the appendix in table \ref{tab:smape_diff_sig} and \ref{tab:mase_diff_sig}, respectively. Aggregate results are summarised in table \ref{tab:summary_replicated}. Our main findings are as follows: 

\begin{table}[htbp]
    \begin{threeparttable}
    \centering \small
    \caption{sMAPE percentage difference between replicated and published results}
    \label{tab:smape_perc_diff}
    \begin{tabular}{@{}lSSSSSS@{}}
\toprule
{} &  {Yearly} &  {Quarterly} &  {Monthly} &  {Weekly} &  {Daily} &  {Hourly} \\
\midrule
Naïve    &     0.000 &        0.000 &     -0.000 &     0.000 &   -0.000 &     0.000 \\
Naïve2   &     0.000 &        0.000 &     -0.000 &     0.000 &   -0.000 &    -0.000 \\
sNaïve   &     0.000 &        0.000 &      0.000 &     0.000 &   -0.000 &     0.000 \\
SES      &    -0.004 &        0.069 &      0.016 &    -0.005 &    0.011 &     0.000 \\
Holt     &     4.063 &       -1.528 &      3.916 &    -3.365 &    0.286 &    -4.347 \\
Damped   &     1.586 &       -1.024 &      0.010 &    -0.694 &    1.036 &    -0.783 \\
Com      &     1.659 &       -0.615 &      1.123 &    -1.418 &    0.498 &    -1.538 \\
ARIMA    &     1.617 &        4.572 &      2.418 &     0.851 &   -2.142 &    -2.017 \\
Theta    &    -1.514 &        0.046 &      0.078 &     0.174 &    0.057 &     0.008 \\
Theta-bc &    -0.948 &       -0.096 &     -0.092 &    -0.043 &    0.050 &     3.378 \\
MLP      &   -11.936 &      -24.109 &    -27.567 &   -52.596 &  -61.727 &    -4.558 \\
RNN      &   -22.728 &      -29.329 &    -30.815 &   -25.979 &  -33.001 &    -9.332 \\
\bottomrule
\end{tabular}
 % load tabular from file
    \begin{tablenotes}
    \setlength{\itemindent}{-2.49997pt} % remove indent from \item below
    \footnotesize
    \item \InsertPercentageDifferenceTableNotes{mean sMAPE}
    \end{tablenotes}
    \end{threeparttable}
\end{table}

\begin{table}[htbp]
    \begin{threeparttable}
    \centering 
    \small
    \caption{Summary of replicated results}
    \label{tab:summary_replicated}
    \begin{tabular}{lrrrrrrrrr}
\toprule
{} & \multicolumn{3}{l}{\textbf{Mean rank (sMAPE)}} & \multicolumn{3}{l}{\textbf{Replicated metrics}} & \multicolumn{3}{l}{\textbf{Running time (min)}} \\
{} &                 Replicated & Original & Change &                       sMAPE &   MASE &    OWA &                  Replicated &  Original & Factor \\
\midrule
Theta-bc &                      5.454 &    5.242 & -0.212 &                      11.952 &  1.583 &  0.876 &                       8.100 &     25.00 &    0.3 \\
Theta    &                      5.649 &    5.437 & -0.212 &                      12.264 &  1.669 &  0.900 &                       6.268 &     12.70 &    0.5 \\
Com      &                      5.726 &    5.454 & -0.271 &                      12.668 &  1.687 &  0.914 &                      69.473 &     33.20 &    2.1 \\
Damped   &                      5.925 &    5.635 & -0.291 &                      12.692 &  1.718 &  0.920 &                      53.448 &     15.30 &    3.5 \\
ARIMA    &                      5.748 &    5.473 & -0.275 &                      12.992 &  1.673 &  0.920 &                   14992.879 &   3030.90 &    4.9 \\
SES      &                      6.788 &    6.700 & -0.088 &                      13.090 &  1.885 &  0.970 &                       5.902 &      8.10 &    0.7 \\
Holt     &                      6.001 &    5.780 & -0.222 &                      14.160 &  1.830 &  0.997 &                      11.720 &     13.30 &    0.9 \\
Naïve2   &                      7.029 &    6.736 & -0.292 &                      13.564 &  1.912 &  1.000 &                       3.664 &      2.90 &    1.3 \\
RNN      &                      6.784 &    8.314 &  1.529 &                      15.122 &  1.902 &  1.067 &                   38941.684 &  64857.10 &    0.6 \\
Naïve    &                      7.337 &    7.050 & -0.287 &                      14.208 &  2.044 &  1.072 &                       1.035 &      0.20 &    5.2 \\
sNaïve   &                      8.046 &    7.729 & -0.316 &                      14.657 &  2.057 &  1.105 &                       1.028 &      0.30 &    3.4 \\
MLP      &                      7.513 &    8.450 &  0.937 &                      16.480 &  2.079 &  1.156 &                     157.884 &   1484.37 &    0.1 \\
\bottomrule
\end{tabular}
 % load tabular from file
    \begin{tablenotes}
    \setlength{\itemindent}{-2.49997pt} % remove indent from \item below
    \footnotesize
    \item \emph{Notes:} Rows show forecasters described in table \ref{tab:m4-models}. Replicated running times are scaled to the number of CPUs used in the original M4 study.  
    \end{tablenotes}
    \end{threeparttable}
\end{table}

\begin{itemize}
    \item For all naïve forecasters, we can replicate the published results perfectly, barring negligible differences due to numerical approximations. This validates our experiment orchestration and evaluation workflow.
    
    \item For the statistical models, we find small but often statistically significant differences. The largest difference we find for sMAPE is $4\%$ for the Holt forecaster on the yearly data set. There appears to be no clear trend in the differences: in some cases, published results are better, in others ours. A possible explanations of the differences is the randomness involved in the optimisation routines used during fitting. However, we do not run the same forecaster multiple times on the same series and hence cannot reliably quantify this source of variation. Differences may also be due to algorithmic differences in the packages we interface and bugs.\footnote{For example, statsmodels' exponential smoothing model seems to return wrong forecasts in a few cases (see  \url{https://github.com/statsmodels/statsmodels/issues/5877}). We also discovered that the M4 study used inconsistent seasonality tests for Python and R which we take into account when replicating the results (see \url{https://github.com/Mcompetitions/M4-methods/issues/25}).}

    \item For MLP and RNN, we find large and statistically significant differences. Differences are entirely negative, ranging from $-11\%$ for MLP on the yearly data set to $-61\%$ on the daily data set. Negative differences indicate that our results are better than those of the M4 study. Again, fitting these models involves randomness, but given the exclusively negative differences, it is likely that the underlying algorithms in scikit-learn and TensorFlow have been improved since the M4 study. This is also suggested by the improved run times shown in table \ref{tab:summary_replicated}. 
    \item In addition, we compare the computational run times between sktime and the M4 study, which for most parts relies on the forecast library in R. Run times are not directly comparable, as we use different machines to replicate the results (see section \ref{sec:technical} for more details). To make run times more comparable, we scale our obtained run times to the number of CPUs used in the M4 study. The scaled values are shown in table \ref{tab:summary_replicated}. Most notably, ARIMA takes approximately $5$x longer than in the M4 study. This is likely because R's forecast library supports the conditional sum of square approximation technique for model estimation \cite[p. 209ff.]{Box1976}, which is considerably faster, especially for long series, but currently not supported by pmdarima and statsmodels. MLP and RNN, based on scikit-learn and TensorFlow, are now substantially faster than in the M4 study. The remaining run times are more or less on par: SES, Holt and Theta are slightly faster when using sktime, the naïve forecasters and Damped slightly slower. 

\end{itemize}

% even under assumption of independence, if difference in performance are not statistically significant, without independence, differences will be even less statistically significant 

%%%%%%%%%%%%%%%%%%%%%%%%%%%%%%%%%%%%%%%%%%%%%%%%%%%%%%%%%%%%%%%%%%%%%%%%%%%%%%%%%%%%
%%%%%%%%%%%%%%%%%%%%%%%%%%%%%%%%%%%%%%%%%%%%%%%%%%%%%%%%%%%%%%%%%%%%%%%%%%%%%%%%%%%%
\subsection{Extending the M4 study}
\label{sec:experiments_extend}

\subsubsection{Research questions}
% intro: purpose & scope
Having replicated key results from the M4 study, we want to extend it for two reasons: First, we want to showcase the usefulness of sktime for solving practical forecasting problems. sktime allows to easily build, tune and evaluate new models thanks to its modular API, including common machine techniques like pipelining, reduction, boosting, and tuning. Second, we want to further investigate the potential of machine learning models for forecasting. In contrast to most of the machine learning entries of the M4 study, we focus on univariate forecasting models that require only a single series during training and hence cannot make use of consistent patterns across multiple series. 

Our extension is guided by three research questions:
\begin{enumerate}
    \item Can standard tabular regression algorithms via reduction outperform statistical models?
    \item Can residual boosting with standard tabular regressors further enhance the predictive performance of Theta-bc, the best  statistical model in the M4 study?
    \item Does tuning the window length hyper-parameter of the reduction from forecasting to tabular regression help improve performance?
    % \item does box-cox help improve performance as in Theta-bc?
    % \item how well do time series regressor perform? 
    % \item how well do deep learning time series regressors perform, given recent success of deep learning on M4 \cite{Oreshkin2019}?
\end{enumerate}

\subsubsection{Models}
To explore these questions, we evaluate five different machine learning approaches. We describe them in detail in table \ref{tab:m4-models-extended}. The simplest approach uses reduction to tabular regression without applying any seasonal adjustments. Instead, we set the window length so that it covers at least a full seasonal period. As in the M4 study, we apply linear detrending in all approaches, as the window slicing of the reduction approach makes it difficult for the models to pick up long-term trends. We evaluate each of the approaches with four standard regression algorithms: Linear regression (LR), K-nearest-neighbours (KNN), random forest (RF), and gradient boosted trees (XGB). For the Theta-based approaches $4$ and $5$, we exclude LR, as Theta already includes some linear extrapolation. This leads to a total of $18$  models that we evaluate in addition to the replicated models. For more details on the regressors and their hyper-parameter settings, see table \ref{tab:regressor} in the appendix. 

\begin{table}[htbp]
    \begin{threeparttable}
    \centering \small
    \caption{Machine learning models}
    \label{tab:m4-models-extended}
    \begin{tabularx}{\textwidth}{lllX}
    \toprule
    \# & Name & Category & Description \\
    \midrule
    1 & \{regressor\} & ML & Regression via reduction, using the standard recursive strategy for generating predictions described in section \ref{sec:reduction}. No seasonal adjustment, but linear detrending and standardisation (removing the mean and scaling to unit variance) is applied. The window length is set to $min(\text{sp}, 3)$, where sp is the seasonal periodicity of the data. \\
    2 & \{regressor\}-s & ML & Like \#1, but with seasonal adjustment as in Naïve2. \\
    3 & \{regressor\}-t-s & ML & Like \#2, but with tuning of the window length. We use a simple temporal cross-validation scheme, in which we make a single split of the training series, using the first window for training and the second window for validation. The validation window has the same length as the forecasting horizon (i.e.\ the test series). We search over the following window length values: $3$, $4$, $6$, $8$, $10$, $12$, $15$, $18$, $21$, $24$. \\
    4 & \{regressor\}-Theta-bc & Hybrid & Residual boosting of Theta-bc. Standardisation is applied as in \#1 to the Theta-bc residuals. Window length is set as in \#1. \\
    5 & \{regressor\}-Theta-bc-t & Hybrid &  Like \#4, but with tuning of the window length as in \#3. \\ 
    \bottomrule
    \end{tabularx}
    \begin{tablenotes}
    \footnotesize
    \item \emph{Notes}: \{regressor\} is a placeholder for the tried out tabular regression algorithms described in the appendix in table \ref{tab:regressor}.
    \end{tablenotes}
    \end{threeparttable}
\end{table}

%%%%%%%%%%%%%%%%%%%%%%%%%%%%%%%%%%%%%%%%%%%%%%%%%%%%%%%%%%%%%%%%%%%%%%%%%%%%%%%%%%%%
\subsubsection{Results}
Detailed results for all models are reported in the appendix in tables \ref{tab:smape_extended}, \ref{tab:mase_extended}, \ref{tab:owa_extended}, and \ref{tab:overall_extended}. Below we discuss our three guiding questions in turn.

\begin{description}
    \item[Question 1:] Can standard tabular regression algorithms via reduction to forecasting achieve equal or better performance than statistical models?
\end{description}

We present selected OWA results in table \ref{tab:extended_ml_models}. We also report some M4 entries as a reference for comparison, including the M4 winner \cite{Smyl2018, Smyl2020}, the runner-up \cite{Montero-Manso2020a}, the best pure machine learning model (a convolutional neural network adapted to time series submitted by Trotta), and Theta-bc as the best statistical forecaster. 

\begin{figure}
    \caption{Critical difference diagram based on sMAPE and the hourly data set}
    \label{fig:cd_q1}
    \centering
    \includegraphics[width=\textwidth]{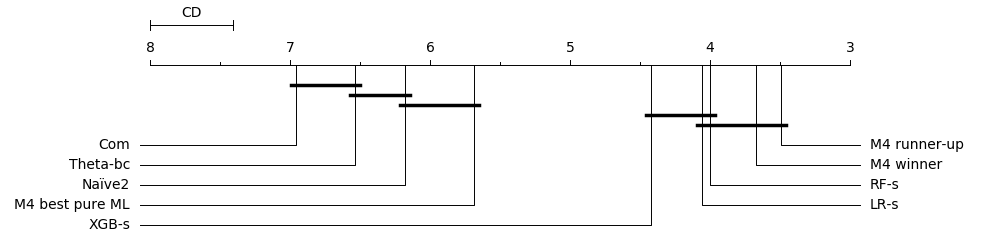}
    \begin{minipage}{\textwidth}
    \footnotesize \emph{Notes:} The diagram is based on sMAPE performance and the hourly data set. On the horizontal line, the diagram shows mean ranks for each forecaster. Forecaster grouped by a bar are not statistically significant based on pairwise post-hoc Nemenyi tests at the $5$\% level. Corresponding Wilcoxon-Holm test results are shown in the appendix in table \ref{tab:wilcoxon_holm_q1}.
    \end{minipage}
\end{figure}

In line with previous results \cite{Makridakis2018, Makridakis2019}, we find that the tried out machine learning models, on average, cannot achieve equal or better performance compared to the  statistical models over the whole of the M4 data set. 

However, on the hourly data set, machine learning models can perform better than statistical ones. Our best model, RF-s with an OWA of $0.493$ comes even close to the multivariate, hybrid models of the M4 winner ($0.44$) and runner-up ($0.484$). XGB-s ($0.496$) and LR-s ($0.501$) achieve slightly worse, but still competitive performances.

To check if found performance differences are statistically significant, we use significance tests and show a critical difference diagram in figure \ref{fig:cd_q1}. Corresponding results from Wilcoxon-Holm tests are shown in the appendix in table \ref{tab:wilcoxon_holm_q1}. For the hourly data set, the differences between RF-s and the statistical models are statistically significant, and that the differences between RF-s and the M4 winner and the runner-up are not statistically significant. 

To answer question 1: No, standard tabular regression algorithms cannot achieve equal or better performance compared to statistical methods on the whole of the M4 data set. But they can achieve equal or better performance on the hourly data set. Indeed, here they even achieve competitive performance compared to the best performing M4 models, with no significant difference between them. 

It is worth emphasising that our tried out models are considerably simpler than the best performing M4 models: they only need a single series for training; they only rely on familiar machine learning techniques; and with sktime, they only require off-the-shelf functionality without any hand-crafted components. 

It also appears that the characteristics of the series may be a critical factor determining the performance of machine learning models. This suggests that certain forecasting problem warrant a more systematic exploration of machine learning models, which we hope to further facilitate with sktime.

\begin{table}[htbp]
    \begin{threeparttable}
    \centering 
    \small
    \caption{Performance of new machine learning models (OWA)}
    \label{tab:extended_ml_models}
    \begin{tabular}{llllllll}
\toprule
{} &           Yearly &        Quarterly &          Monthly &           Weekly &            Daily &          Hourly &            Total \\
\midrule
M4 winner       &            0.778 &  \bfseries 0.847 &  \bfseries 0.836 &            0.851 &            1.046 &  \bfseries 0.44 &  \bfseries 0.833 \\
M4 runner-up    &            0.799 &  \bfseries 0.847 &            0.858 &  \bfseries 0.796 &            1.019 &           0.484 &            0.847 \\
Theta-bc        &  \bfseries 0.776 &            0.893 &            0.904 &            0.964 &            0.996 &           1.009 &            0.876 \\
Com             &            0.886 &            0.885 &             0.93 &            0.911 &  \bfseries 0.982 &           1.506 &            0.914 \\
M4 best pure ML &            0.859 &            0.939 &            0.941 &            0.996 &            1.071 &           0.634 &            0.926 \\
RF-s            &            0.967 &            1.014 &            0.994 &            1.015 &            1.078 &           0.493 &            0.994 \\
Naïve2          &              1.0 &              1.0 &              1.0 &              1.0 &              1.0 &             1.0 &              1.0 \\
XGB-s           &            1.022 &            1.091 &            1.118 &            1.113 &            1.149 &           0.496 &            1.088 \\
LR-s            &                - &            1.037 &             2.16 &            0.964 &             1.07 &           0.501 &                - \\
\bottomrule
\end{tabular}

    \begin{tablenotes}
    \setlength{\itemindent}{-2.49997pt} % remove indent from \item below
    \footnotesize
    \item \emph{Notes:} Rows show forecasters described in table \ref{tab:m4-models-extended}. Columns show M4 data sets grouped by sampling frequency. We exclude results for LR model when generated forecasts were instable/exploding due the little available data and linear extrapolation.
    \end{tablenotes}
    \end{threeparttable}
\end{table}

%%%%%%%%%%%%%%%%%%%%%%%%%%%%%%%%%%%%%%%%%%%%%%%%%%%%%%%%%%%%%%%%%%%%%%%%%%%%%%%%%%%%
\begin{description}
    \item[Question 2:] Can residual boosting with standard tabular regressors further enhance the predictive performance of Theta-bc, the best  statistical model in the M4 study?
\end{description}

To explore the second question, we compare results for Theta-bc with their boosted variants based on the RF and XGB regression algorithms. 
%While this approach is inspired by the hybrid M4 winning hybrid model \cite{Smyl2018, Smyl2020}, it is considerably simpler. 
We also include the tuned versions of the boosted models. Results are shown in table \ref{tab:extended_theta_boosting}. Our key findings are as follows:

On the whole of the M4 data set, boosting does not improve the performance of Theta-bc. While boosting leads to a performance loss on the yearly, monthly and quarterly data sets, it leads to slight performance gains on the weekly, daily and hourly data set. In particular, on the daily data set, boosting with RF improves the OWA of Theta-bc from $0.996$ to $0.988$, and on the hourly data set from $1.009$ to $0.985$. For the weekly data set, tuning is needed to improve accuracy beyond that of Theta-bc. 

\begin{figure}
    \caption{Critical difference diagram based on sMAPE and the hourly data set}
    \label{fig:cd_q2}
    \centering
    \includegraphics[width=\textwidth]{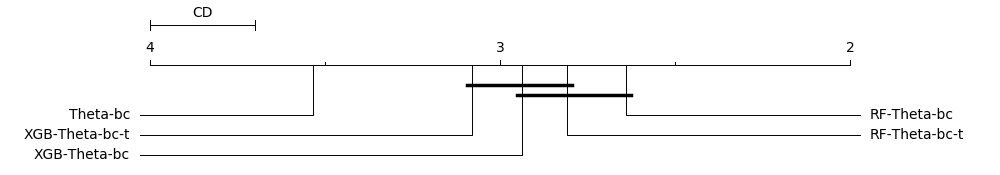}
    \begin{minipage}{\textwidth}
    \footnotesize \emph{Notes:} The diagram is based on sMAPE performance and the hourly data set. On the horizontal line, the diagram shows mean ranks for each forecaster. Forecaster grouped by a bar are not statistically significant based on pairwise post-hoc Nemenyi tests at the $5$\% level. Corresponding Wilcoxon-Holm test results are shown in the appendix in table \ref{tab:wilcoxon_holm_q2}.
    \end{minipage}
\end{figure}

Again, we test whether found performance differences on the hourly data set are significant. As shown in figure \ref{fig:cd_q2}, we find that the boosted variants perform significantly better than Theta-bc without boosting, with no significant difference between the boosted variants. 

\begin{table}[htbp]
    \begin{threeparttable}
    \centering 
    \small
    \caption{Performance of boosted Theta-bc models (OWA)}
    \label{tab:extended_theta_boosting}
    \begin{tabular}{llllllll}
\toprule
{} &           Yearly &        Quarterly &          Monthly &           Weekly &            Daily &           Hourly &            Total \\
\midrule
Theta-bc       &  \bfseries 0.776 &  \bfseries 0.893 &  \bfseries 0.904 &            0.964 &            0.996 &            1.009 &  \bfseries 0.876 \\
RF-Theta-bc-t  &            0.854 &            0.938 &            0.933 &  \bfseries 0.959 &            0.999 &            0.987 &            0.919 \\
RF-Theta-bc    &            0.864 &            0.957 &            0.932 &            1.052 &  \bfseries 0.988 &  \bfseries 0.985 &            0.925 \\
XGB-Theta-bc   &            0.898 &            1.017 &            0.971 &            1.153 &            1.023 &            0.993 &            0.968 \\
XGB-Theta-bc-t &            0.906 &            1.009 &            0.976 &            1.006 &             1.04 &            0.998 &            0.971 \\
\bottomrule
\end{tabular}

    \begin{tablenotes}
    \setlength{\itemindent}{-2.49997pt} % remove indent from \item below
    \footnotesize
    \item \emph{Notes:} Rows show forecasters described in table \ref{tab:m4-models-extended}. Columns show M4 data sets grouped by sampling frequency.
    \end{tablenotes}
    \end{threeparttable}
\end{table}

%%%%%%%%%%%%%%%%%%%%%%%%%%%%%%%%%%%%%%%%%%%%%%%%%%%%%%%%%%%%%%%%%%%%%%%%%%%%%%%%%%%%
\begin{description}
    \item[Question 3:] Does tuning the window length hyper-parameter of the reduction from forecasting to tabular regression help improve performance?
\end{description}

To explore the last question, we compare the performance of each regressor, once with a tuned window length and once with the default window length (see table \ref{tab:m4-models-extended}). Results are shown in table \ref{tab:extended_tuning}. Our key findings are as follows:

In general, tuning of the window length does not help improve performance. Exceptions are the weekly data set and the KNN regressor. On the weekly data set, all tried out regressors benefit from tuning, with the biggest OWA improvement being $0.082$ for XGB. KNN additionally benefits from tuning on the yearly and hourly data set. 

Note that other temporal cross-validation schemes than the one tried out here are possible and may prove to be more beneficial to overall performance (e.g.\ a sliding window validation). In addition, we may want to try to optimise additional hyper-parameters (e.g.\ the strategy to generate forecasts as discussed in \ref{sec:reduction}, or the hyper-parameters of regressor using a nested tabular cross-validation scheme). But note that tuning comes at the cost of a considerable increase in computational running time, as reported in the appendix in table \ref{tab:overall_extended}. 

\begin{table}[htbp]
    \begin{threeparttable}
    \centering 
    \small
    \caption{Performance of tuned machine learning models (OWA)}
    \label{tab:extended_tuning}
    \begin{tabular}{lrrrrrrr}
\toprule
{} &  Yearly &  Quarterly &  Monthly &  Weekly &  Daily &  Hourly &  Total \\
\midrule
RF-s    &   0.967 &      1.014 &    0.994 &   1.015 &  1.078 &   0.493 &  0.994 \\
RF-t-s  &   1.005 &      1.070 &    1.057 &   0.967 &  1.087 &   0.683 &  1.047 \\
XGB-s   &   1.022 &      1.091 &    1.118 &   1.113 &  1.149 &   0.496 &  1.088 \\
XGB-t-s &   1.038 &      1.144 &    1.170 &   1.031 &  1.179 &   0.746 &  1.131 \\
KNN-s   &   1.086 &      1.171 &    1.257 &   1.218 &  1.338 &   0.544 &  1.197 \\
KNN-t-s &   1.062 &      1.185 &    1.276 &   1.147 &  1.331 &   0.751 &  1.205 \\
LR-s    &       - &      1.037 &    2.160 &   0.964 &  1.070 &   0.501 &      - \\
LR-t-s  &       - &          - &    1.876 &   0.889 &  1.089 &   0.670 &      - \\
\bottomrule
\end{tabular}

    \begin{tablenotes}
    \setlength{\itemindent}{-2.49997pt} % remove indent from \item below
    \footnotesize
    \item \emph{Notes:} Rows show forecasters described in table \ref{tab:m4-models-extended}. Columns show M4 data sets grouped by sampling frequency. \InsertLRTableNotes
    \end{tablenotes}
    \end{threeparttable}
\end{table}

%%%%%%%%%%%%%%%%%%%%%%%%%%%%%%%%%%%%%%%%%%%%%%%%%%%%%%%%%%%%%%%%%%%%%%%%%%%%%%%%%%%%
%%%%%%%%%%%%%%%%%%%%%%%%%%%%%%%%%%%%%%%%%%%%%%%%%%%%%%%%%%%%%%%%%%%%%%%%%%%%%%%%%%%%
\section{Conclusion}
\label{sec:conclusion}
We presented sktime's new forecasting framework. sktime integrates with scikit-learn as one of the major machine learning toolboxes, and allows to easily build, tune and evaluate composite machine learning models. We discussed key features of sktime's forecasting API, including composite models familiar from scikit-learn, but also novel meta-forecasters for reduction and detrending. 

In addition, we replicated and extended the M4 forecasting study. Replicating the M4 study allowed us to test our model implementations, and validate  published results. We found no or small differences for the naïve and statistical forecasting algorithms, and larger improvements for the machine learning algorithms. 

Extending the M4 study allowed us to highlight the usefulness of sktime and to further investigate the potential of simple off-the-shelf machine learning approaches for univariate forecasting. In particular, we found that simple pure approaches like  reduction, pipelining and tuning can achieve competitive forecasting performance on the hourly data set, outperforming the statistical algorithms and coming close to the best M4 models. In addition, we found that simple hybrid approaches using residual boosting of statistical methods can help improve their forecasting performance in some cases.

With sktime, we hope to further advance the available toolbox capabilities for time series analysis. In future work, we want to further develop sktime by adding full support for:
\begin{itemize}
    \item Time series regression algorithms, refactoring existing time series classification algorithm as well as adding bespoke time series regressors, 
    \item Exogenous, multivariate time series, extending bespoke algorithms and adding modular composition techniques specifically for multivariate series, 
    \item Prediction intervals and probabilistic forecasting.
\end{itemize}
In addition, we hope to develop new frameworks for related learning tasks, including multivariate/panel forecasting and time series annotation (e.g.\ segmentation and outlier detection). 

%%%%%%%%%%%%%%%%%%%%%%%%%%%%%%%%%%%%%%%%%%%%%%%%%%%%%%%%%%%%%%%%%%%%%%%%%%%%%%%%%%%%
%%%%%%%%%%%%%%%%%%%%%%%%%%%%%%%%%%%%%%%%%%%%%%%%%%%%%%%%%%%%%%%%%%%%%%%%%%%%%%%%%%%%
\newpage
\section*{Acknowledgements}
The first phase of development for sktime was done jointly between researchers at the University of East Anglia (UEA), University College London (UCL) and The Alan Turing Institute as part of a UK Research and Innovation (UKRI) project to develop tools for data science and artificial intelligence.

We want to thank all contributors on GitHub, with a special thanks to @big-o (GitHub username) for the Python implementation of the Theta forecaster, and @matteogales, @big-o, and Patrick Rockenschaub for feedback on the interface design. 

We are also grateful to the M4 organising team for their support to replicate the results of the M4 study. 

Markus Löning's contribution was supported by the Economic and Social Research Council (ESRC) [grant: ES/P000592/1], the Consumer Data Research Centre (CDRC) [ESRC grant: ES/L011840/1], and The Alan Turing Institute (EPSRC grant no. EP/N510129/1).

%%%%%%%%%%%%%%%%%%%%%%%%%%%%%%%%%%%%%%%%%%%%%%%%%%%%%%%%%%%%%%%%%%%%%%%%%%%%%%%%%%%%
%%%%%%%%%%%%%%%%%%%%%%%%%%%%%%%%%%%%%%%%%%%%%%%%%%%%%%%%%%%%%%%%%%%%%%%%%%%%%%%%%%%%
\section*{Authors' contributions}
ML made key contributions to architecture and design, including composition and reduction interfaces. ML is one of sktime's lead developers, having contributed to almost all parts of it, including the overall toolbox architecture, the time series classification framework, and specific algorithms. ML drafted and wrote most of this manuscript. He wrote the code to replicate and extend the M4 study. 

FK conceived the project and architectural outlines, including taxonomy of learning tasks, composition approaches and reduction relations. FK further made key contributions to architecture and design, and contributed to writing of this manuscript.

%%%%%%%%%%%%%%%%%%%%%%%%%%%%%%%%%%%%%%%%%%%%%%%%%%%%%%%%%%%%%%%%%%%%%%%%%%%%%%%%%%%%
%%%%%%%%%%%%%%%%%%%%%%%%%%%%%%%%%%%%%%%%%%%%%%%%%%%%%%%%%%%%%%%%%%%%%%%%%%%%%%%%%%%%
%\section*{References}
\newpage
\bibliographystyle{unsrtnat}
\bibliography{references}

%%%%%%%%%%%%%%%%%%%%%%%%%%%%%%%%%%%%%%%%%%%%%%%%%%%%%%%%%%%%%%%%%%%%%%%%%%%%%%%%%%%%
%%%%%%%%%%%%%%%%%%%%%%%%%%%%%%%%%%%%%%%%%%%%%%%%%%%%%%%%%%%%%%%%%%%%%%%%%%%%%%%%%%%%
\newpage
\section*{Appendix}
\appendix

%%%%%%%%%%%%%%%%%%%%%%%%%%%%%%%%%%%%%%%%%%%%%%%%%%%%%%%%%%%%%%%%%%%%%%%%%%%%%%%%%%%%
\begin{table}[htbp]
\begin{threeparttable}
\centering \small
\caption{The number of M4 series per sampling frequency and domain}
\label{tab:m4-domain}
\begin{tabular}{lrrrrrrr}
\toprule
{} &  Demographic &  Finance &  Industry &  Macro &  Micro &  Other &   Total \\
\midrule
Yearly    &         1088 &     6519 &      3716 &   3903 &   6538 &   1236 &   23000 \\
Quarterly &         1858 &     5305 &      4637 &   5315 &   6020 &    865 &   24000 \\
Monthly   &         5728 &    10987 &     10017 &  10016 &  10975 &    277 &   48000 \\
Weekly    &           24 &      164 &         6 &     41 &    112 &     12 &     359 \\
Daily     &           10 &     1559 &       422 &    127 &   1476 &    633 &    4227 \\
Hourly    &            0 &        0 &         0 &      0 &      0 &    414 &     414 \\
Total     &         8708 &    24534 &     18798 &  19402 &  25121 &   3437 &  100000 \\
\bottomrule
\end{tabular}

\begin{tablenotes}
\setlength{\itemindent}{-2.49997pt}
\footnotesize
\item \emph{Notes}: Rows show M4 data sets grouped by sampling frequency. Columns show M4 data sets grouped by domains. Values show the number of available series.   
\end{tablenotes}
\end{threeparttable}
\end{table}

%%%%%%%%%%%%%%%%%%%%%%%%%%%%%%%%%%%%%%%%%%%%%%%%%%%%%%%%%%%%%%%%%%%%%%%%%%%%%%%%%%%%
\begin{table}[htbp]
\begin{threeparttable}
\centering \small
\caption{Summary statistics of the length of time series in the training set}
\label{tab:m4-training}
\begin{tabular}{lrrrrrrr}
\toprule
{} &    Mean &     Std &  Min &  25\% &   50\% &   75\% &   Max \\
\midrule
Yearly    &    31.3 &    24.5 &   13 &   20 &    29 &    40 &   835 \\
Quarterly &    92.3 &    51.1 &   16 &   62 &    88 &   115 &   866 \\
Monthly   &   216.3 &   137.4 &   42 &   82 &   202 &   306 &  2794 \\
Weekly    &  1022.0 &   707.1 &   80 &  379 &   934 &  1603 &  2597 \\
Daily     &  2357.4 &  1756.6 &   93 &  323 &  2940 &  4197 &  9919 \\
Hourly    &   853.9 &   127.9 &  700 &  700 &   960 &   960 &   960 \\
Total     &   240.0 &   592.3 &   13 &   49 &    97 &   234 &  9919 \\
\bottomrule
\end{tabular}

\begin{tablenotes}
\setlength{\itemindent}{-2.49997pt}
\footnotesize
\item \emph{Notes}: Rows show M4 data sets grouped by sampling frequency. Columns show summary statistics of the distribution of the length of the training series.
\end{tablenotes}
\end{threeparttable}
\end{table}

%%%%%%%%%%%%%%%%%%%%%%%%%%%%%%%%%%%%%%%%%%%%%%%%%%%%%%%%%%%%%%%%%%%%%%%%%%%%%%%%%%%%
\begin{table}[htbp]
    \begin{threeparttable}
    \centering \small
    \caption{Replicated M4 forecasters}
    \label{tab:m4-models}
    \begin{tabularx}{\textwidth}{llX}
    \toprule
         Name & Category & Description \\
    \midrule
         Naïve & naïve & Always predicting the last observed values. \\
         sNaïve & naïve & Always predicting the last observed value of the same season. \\
         Naïve2 & naïve & Like Naïve, but with seasonal adjustment by applying classical multiplicative decomposition \cite{Cleveland1990} and an autocorrelation test at the 90\% significance level to decide whether or not to apply seasonal adjustment \cite{Fiorucci2016}. \\
         SES & statistical & Simple exponential smoothing and extrapolating \cite{Holt1957, Winters1960, Holt2004a}, with no trend, and seasonal adjustment as in Naïve2.\\
         Holt & statistical & Like SES, but with linear trend. \\
         Damped & statistical & Like Holt, but with damped trend \cite{Gardner1985}. \\
         Theta & statistical & As applied to the M3 Competition \cite{Makridakis2000} using two Theta lines, $\theta_1=0$ and $\theta_2=2$, with the first one being extrapolated using linear regression and the second one using SES. The forecasts are then combined using equal weights \cite{Assimakopoulos2000a}. This is equivalent to special case of simple exponential smoothing with drift \cite{Hyndman2003}. Seasonal adjustments are considered as in Naïve2.\\
         Theta-bc & statistical & Like Theta, but with Box-Cox adjustment \cite{Box1964}, where lambda is constraint to the $(0, 1)$ interval and estimated via maximum likelihood estimation. Submitted to the M4 study by Legaki \& Koutsouri (submission number 260). \\
         Com & statistical & Simple arithmetic mean of SES, Holt and Damped.\\
         ARIMA & statistical & An automatic selection of possible seasonal ARIMA models is performed and the best one is chosen using appropriate selection criteria \cite{Box2015, Hyndman2008}. \\
         MLP & ML & A multi-layer perceptron of a very basic architecture and parameterization via a recursive reduction approach as described in section \ref{sec:reduction} with window length set to $3$. Linear detrending and seasonal adjustments as in Naïve2 is applied to facilitate extrapolation.\\
         RNN & ML & A recurrent network of a very basic architecture and parameterization via a recursive reduction approach as described in section \ref{sec:reduction} with window length set to $3$. Linear detrending and seasonal adjustments as in Naïve2 is applied to facilitate extrapolation. \\
    \bottomrule
    \end{tabularx}
    \begin{tablenotes}
    \setlength{\itemindent}{-2.49997pt}
    \footnotesize
    \item \emph{Notes}: The forecasters are described in detail in the original M4 study \cite{Makridakis2019}. We follow the categorisation of the M4 study here for consistency, but more fruitful categorisations have been proposed by \cite{Januschowski2020}. 
    \end{tablenotes}
    \end{threeparttable}
\end{table}

%%%%%%%%%%%%%%%%%%%%%%%%%%%%%%%%%%%%%%%%%%%%%%%%%%%%%%%%%%%%%%%%%%%%%%%%%%%%%%%%%%%%

\begin{table}[htbp]
    \begin{threeparttable}
    \centering \small
    \caption{MASE percentage difference between published and replicated results}
    \label{tab:mase_perc_diff}
    \sisetup{
      table-align-uncertainty=true,
      separate-uncertainty=true,
    }
    \begin{tabular}{@{}lSSSSSS@{}}
\toprule
{} &  {Yearly} &  {Quarterly} &  {Monthly} &  {Weekly} &  {Daily} &  {Hourly} \\
\midrule
Naïve    &     0.000 &        0.000 &      0.000 &     0.000 &    0.000 &     0.000 \\
Naïve2   &     0.000 &        0.000 &      0.000 &     0.000 &    0.000 &    -0.000 \\
sNaïve   &     0.000 &        0.000 &      0.000 &     0.000 &    0.000 &     0.000 \\
SES      &    -0.008 &        0.015 &      0.038 &    -0.002 &    0.004 &     0.000 \\
Holt     &     5.586 &        0.001 &      2.805 &    -1.599 &    0.200 &    -3.927 \\
Damped   &     3.696 &       -0.458 &      1.507 &    -2.887 &    0.814 &    -1.154 \\
Com      &     2.751 &       -0.425 &      1.035 &    -1.900 &    0.334 &    -4.332 \\
ARIMA    &     0.141 &        3.329 &      0.991 &   -11.172 &   -4.873 &     4.054 \\
Theta    &    -3.058 &       -1.153 &     -0.139 &    -0.016 &   -0.041 &     0.095 \\
Theta-bc &    -1.922 &       -1.013 &     -0.202 &    -0.723 &   -0.035 &    -3.606 \\
MLP      &   -14.658 &      -30.207 &    -41.370 &   -80.142 &  -70.993 &    -9.859 \\
RNN      &   -24.522 &      -34.106 &    -30.284 &   -38.831 &  -36.420 &   -20.987 \\
\bottomrule
\end{tabular}
 % load tabular from file
    \begin{tablenotes}
    \setlength{\itemindent}{-2.49997pt} % remove indent from \item below
    \footnotesize
    \item \InsertPercentageDifferenceTableNotes{MASE}
    \end{tablenotes}
    \end{threeparttable}
\end{table}

%%%%%%%%%%%%%%%%%%%%%%%%%%%%%%%%%%%%%%%%%%%%%%%%%%%%%%%%%%%%%%%%%%%%%%%%%%%%%%%%%%%%

\begin{table}[htbp]
\begin{threeparttable}
\centering \small
\caption{OWA percentage difference between replicated and published results}
\label{tab:owa_perc_diff}
\begin{tabular}{@{}lSSSSSS@{}}
\toprule
{} &  {Yearly} &  {Quarterly} &  {Monthly} &  {Weekly} &  {Daily} &  {Hourly} \\
\midrule
Naïve    &     0.000 &        0.000 &      0.000 &     0.000 &    0.000 &     0.000 \\
Naïve2   &     0.000 &        0.000 &      0.000 &     0.000 &    0.000 &     0.000 \\
sNaïve   &     0.000 &       -0.000 &      0.000 &     0.000 &    0.000 &     0.000 \\
SES      &    -0.006 &        0.042 &      0.027 &    -0.003 &    0.007 &     0.000 \\
Holt     &     4.782 &       -0.812 &      3.382 &    -2.568 &    0.244 &    -4.048 \\
Damped   &     2.594 &       -0.753 &      0.751 &    -1.729 &    0.926 &    -0.984 \\
Com      &     2.179 &       -0.524 &      1.080 &    -1.646 &    0.416 &    -3.256 \\
ARIMA    &     0.909 &        3.984 &      1.727 &    -5.082 &   -3.502 &     0.053 \\
Theta    &    -2.267 &       -0.541 &     -0.031 &     0.081 &    0.008 &     0.053 \\
Theta-bc &    -1.416 &       -0.542 &     -0.147 &    -0.372 &    0.008 &    -0.308 \\
MLP      &   -13.251 &      -27.165 &    -34.713 &   -71.246 &  -66.951 &    -7.691 \\
RNN      &   -23.582 &      -31.657 &    -30.563 &   -32.746 &  -34.685 &   -16.490 \\
\bottomrule
\end{tabular}
 % load tabular from file
\begin{tablenotes}
\setlength{\itemindent}{-2.49997pt} % remove indent from \item below
\footnotesize
\item \InsertPercentageDifferenceTableNotes{OWA}
\end{tablenotes}
\end{threeparttable}
\end{table}

%%%%%%%%%%%%%%%%%%%%%%%%%%%%%%%%%%%%%%%%%%%%%%%%%%%%%%%%%%%%%%%%%%%%%%%%%%%%%%%%%%%%

\begin{table}[htbp]
    \begin{threeparttable}
    \centering \small
    \caption{sMAPE difference between published and replicated results}
    \label{tab:smape_diff_sig}
    \sisetup{
      table-align-uncertainty=true,
      separate-uncertainty=true,
    }
    \begin{tabular}{@{}l*{6}{S[table-format=2.3(3),mode=text,detect-weight=true]}@{}}
\toprule
{} &                    {Yearly} &                 {Quarterly} &                   {Monthly} &                     {Weekly} &                     {Daily} &                    {Hourly} \\
\midrule
Naïve    &                -0.0 \pm 0.0 &                 0.0 \pm 0.0 &                 0.0 \pm 0.0 &                  0.0 \pm 0.0 &                 0.0 \pm 0.0 &                 0.0 \pm 0.0 \\
Naïve2   &                -0.0 \pm 0.0 &                -0.0 \pm 0.0 &                 0.0 \pm 0.0 &                  0.0 \pm 0.0 &                 0.0 \pm 0.0 &                 0.0 \pm 0.0 \\
sNaïve   &                -0.0 \pm 0.0 &                 0.0 \pm 0.0 &                 0.0 \pm 0.0 &                  0.0 \pm 0.0 &                 0.0 \pm 0.0 &                 0.0 \pm 0.0 \\
SES      &            -0.001 \pm 0.006 &             0.007 \pm 0.004 &             0.002 \pm 0.001 &                 -0.0 \pm 0.0 &                 0.0 \pm 0.0 &                 0.0 \pm 0.0 \\
Holt     &   \bfseries 0.665 \pm 0.083 &  \bfseries -0.167 \pm 0.034 &    \bfseries 0.58 \pm 0.044 &   \bfseries -0.327 \pm 0.107 &             0.009 \pm 0.013 &  \bfseries -1.271 \pm 0.463 \\
Damped   &   \bfseries 0.241 \pm 0.056 &  \bfseries -0.105 \pm 0.023 &              0.001 \pm 0.02 &             -0.062 \pm 0.042 &             0.032 \pm 0.021 &            -0.151 \pm 0.151 \\
Com      &   \bfseries 0.246 \pm 0.042 &  \bfseries -0.063 \pm 0.016 &   \bfseries 0.151 \pm 0.016 &   \bfseries -0.127 \pm 0.037 &   \bfseries 0.015 \pm 0.007 &  \bfseries -0.339 \pm 0.171 \\
ARIMA    &   \bfseries 0.245 \pm 0.058 &   \bfseries 0.477 \pm 0.038 &   \bfseries 0.325 \pm 0.035 &              0.074 \pm 0.285 &  \bfseries -0.068 \pm 0.023 &            -0.282 \pm 0.389 \\
Theta    &  \bfseries -0.221 \pm 0.017 &             0.005 \pm 0.007 &    \bfseries 0.01 \pm 0.004 &    \bfseries 0.016 \pm 0.006 &             0.002 \pm 0.002 &     \bfseries 0.001 \pm 0.0 \\
Theta-bc &  \bfseries -0.127 \pm 0.017 &             -0.01 \pm 0.005 &  \bfseries -0.012 \pm 0.004 &             -0.004 \pm 0.006 &             0.002 \pm 0.001 &   \bfseries 0.593 \pm 0.242 \\
MLP      &  \bfseries -2.598 \pm 0.048 &   \bfseries -4.46 \pm 0.062 &  \bfseries -6.708 \pm 0.062 &  \bfseries -11.229 \pm 1.046 &  \bfseries -5.754 \pm 0.179 &  \bfseries -0.631 \pm 0.127 \\
RNN      &  \bfseries -5.091 \pm 0.091 &  \bfseries -4.994 \pm 0.073 &  \bfseries -7.413 \pm 0.089 &   \bfseries -3.954 \pm 0.785 &  \bfseries -1.968 \pm 0.113 &  \bfseries -1.372 \pm 0.383 \\
\bottomrule
\end{tabular}
 % load tabular from file
    \begin{tablenotes}
    \setlength{\itemindent}{-2.49997pt} % remove indent from \item below
    \footnotesize
    \item \InsertDifferenceTableNotes{sMAPE}
    \end{tablenotes}
    \end{threeparttable}
\end{table}

%%%%%%%%%%%%%%%%%%%%%%%%%%%%%%%%%%%%%%%%%%%%%%%%%%%%%%%%%%%%%%%%%%%%%%%%%%%%%%%%%%%%

\begin{table}[htbp]
    \begin{threeparttable}
    \centering \small
    \caption{MASE difference between published and replicated results}
    \label{tab:mase_diff_sig}
    \sisetup{
      table-align-uncertainty=true,
      separate-uncertainty=true,
    }
    \begin{tabular}{@{}l*{6}{S[table-format=2.3(3),mode=text,detect-weight=true]}@{}}
\toprule
{} &                    {Yearly} &                 {Quarterly} &                   {Monthly} &                     {Weekly} &                     {Daily} &                    {Hourly} \\
\midrule
Naïve    &                 0.0 \pm 0.0 &                 0.0 \pm 0.0 &                 0.0 \pm 0.0 &                  0.0 \pm 0.0 &                 0.0 \pm 0.0 &                 0.0 \pm 0.0 \\
Naïve2   &                 0.0 \pm 0.0 &                 0.0 \pm 0.0 &                 0.0 \pm 0.0 &                  0.0 \pm 0.0 &                 0.0 \pm 0.0 &                 0.0 \pm 0.0 \\
sNaïve   &                 0.0 \pm 0.0 &                 0.0 \pm 0.0 &                 0.0 \pm 0.0 &                  0.0 \pm 0.0 &                 0.0 \pm 0.0 &                 0.0 \pm 0.0 \\
SES      &                -0.0 \pm 0.0 &                 0.0 \pm 0.0 &       \bfseries 0.0 \pm 0.0 &                 -0.0 \pm 0.0 &                 0.0 \pm 0.0 &       \bfseries 0.0 \pm 0.0 \\
Holt     &   \bfseries 0.198 \pm 0.012 &               0.0 \pm 0.002 &   \bfseries 0.028 \pm 0.002 &   \bfseries -0.039 \pm 0.012 &             0.006 \pm 0.009 &  \bfseries -0.367 \pm 0.168 \\
Damped   &   \bfseries 0.125 \pm 0.008 &  \bfseries -0.005 \pm 0.002 &   \bfseries 0.015 \pm 0.005 &   \bfseries -0.069 \pm 0.023 &             0.026 \pm 0.024 &            -0.034 \pm 0.116 \\
Com      &    \bfseries 0.09 \pm 0.006 &  \bfseries -0.005 \pm 0.001 &    \bfseries 0.01 \pm 0.002 &   \bfseries -0.046 \pm 0.012 &             0.011 \pm 0.007 &  \bfseries -0.199 \pm 0.062 \\
ARIMA    &              0.005 \pm 0.01 &   \bfseries 0.039 \pm 0.003 &   \bfseries 0.009 \pm 0.002 &   \bfseries -0.286 \pm 0.115 &  \bfseries -0.166 \pm 0.024 &             0.038 \pm 0.024 \\
Theta    &  \bfseries -0.103 \pm 0.002 &    \bfseries -0.014 \pm 0.0 &    \bfseries -0.001 \pm 0.0 &               -0.0 \pm 0.003 &  \bfseries -0.001 \pm 0.001 &     \bfseries 0.002 \pm 0.0 \\
Theta-bc &  \bfseries -0.058 \pm 0.005 &  \bfseries -0.012 \pm 0.001 &    \bfseries -0.002 \pm 0.0 &   \bfseries -0.019 \pm 0.008 &            -0.001 \pm 0.001 &             -0.092 \pm 0.05 \\
MLP      &  \bfseries -0.725 \pm 0.027 &  \bfseries -0.699 \pm 0.019 &  \bfseries -0.796 \pm 0.045 &  \bfseries -10.874 \pm 5.053 &   \bfseries -9.21 \pm 1.916 &  \bfseries -0.257 \pm 0.033 \\
RNN      &  \bfseries -1.213 \pm 0.028 &  \bfseries -0.688 \pm 0.008 &  \bfseries -0.485 \pm 0.005 &   \bfseries -1.993 \pm 0.263 &   \bfseries -2.27 \pm 0.114 &   \bfseries -0.64 \pm 0.248 \\
\bottomrule
\end{tabular}
 % load tabular from file
    \begin{tablenotes}
    \setlength{\itemindent}{-2.49997pt} % remove indent from \item below
    \footnotesize
    \item \InsertDifferenceTableNotes{MASE}
    \end{tablenotes}
    \end{threeparttable}
\end{table}

%%%%%%%%%%%%%%%%%%%%%%%%%%%%%%%%%%%%%%%%%%%%%%%%%%%%%%%%%%%%%%%%%%%%%%%%%%%%%%%%%%%%
\begin{table}[htbp]
    \begin{threeparttable}
    \centering \small
    \caption{Tabular regression algorithms used to extend the M4 study}
    \label{tab:regressor}
    \begin{tabular}{lll}
    \toprule
    Name & Description & Hyper-parameters \\
    \midrule
    LR & Linear regression & fit\_intercept=True \\
    % Ridge & Ridge regression &\\
    KNN & K-nearest neighbours & n\_neighbors=1\\
    % SVR & Support vector regression &\\
    RF & Random forest & n\_estimators=500 \\
    XGB & Gradient boosted trees & n\_estimators=500 \\
    % MLPx & Multi-layer perceptron & \\
    \bottomrule
    \end{tabular}
    \begin{tablenotes}
    \footnotesize
    \item \emph{Notes}: For all algorithms except XGB, we use scikit-learn. For XGB, we use xgboost \cite{Chen2016}. For all other hyper-parameters, we use the packages' default settings. 
    \end{tablenotes}
    \end{threeparttable}
\end{table}

%%%%%%%%%%%%%%%%%%%%%%%%%%%%%%%%%%%%%%%%%%%%%%%%%%%%%%%%%%%%%%%%%%%%%%%%%%%%%%%%%%%%
\begin{table}[htbp]
    \begin{threeparttable}
    \centering \small
    \caption{Complete results (sMAPE)}
    \label{tab:smape_extended}
    \sisetup{
      table-align-uncertainty=true,
      separate-uncertainty=true,
    }
    \begin{tabular}{lllllll}
\toprule
{} &            Yearly &         Quarterly &          Monthly &           Weekly &            Daily &            Hourly \\
\midrule
Theta-bc       &  \bfseries 13.239 &            10.145 &  \bfseries 12.99 &            9.144 &            3.042 &             18.16 \\
Theta          &            14.372 &            10.316 &           13.012 &            9.109 &            3.055 &             18.14 \\
Com            &            15.094 &  \bfseries 10.112 &           13.585 &            8.817 &  \bfseries 2.995 &            21.714 \\
RF-Theta-bc-t  &            14.552 &            10.788 &           13.543 &            9.014 &            3.056 &            18.136 \\
Damped         &            15.439 &            10.132 &           13.475 &            8.804 &            3.096 &            19.114 \\
RF-Theta-bc    &            14.738 &            11.006 &           13.527 &            9.801 &            3.061 &            18.061 \\
ARIMA          &            15.413 &            10.908 &           13.768 &            8.727 &            3.124 &            13.698 \\
SES            &            16.395 &            10.607 &            13.62 &            9.011 &            3.045 &            18.094 \\
XGB-Theta-bc-t &            15.334 &            11.535 &           14.166 &            9.484 &            3.134 &            18.303 \\
XGB-Theta-bc   &            15.395 &            11.651 &           14.109 &           10.808 &             3.16 &            18.199 \\
Naïve2         &            16.342 &            11.012 &           14.427 &            9.161 &            3.045 &            18.383 \\
KNN-Theta-bc   &            15.503 &            12.046 &           14.805 &           11.133 &            3.189 &             19.32 \\
KNN-Theta-bc-t &            15.507 &            12.036 &           14.877 &            10.17 &            3.189 &            19.361 \\
RF-s           &            16.655 &            11.674 &           14.899 &            9.327 &            3.291 &            10.978 \\
RF             &            16.656 &            11.793 &            15.01 &             9.31 &            3.289 &            13.663 \\
Holt           &            17.018 &             10.74 &           15.393 &            9.381 &            3.075 &            27.978 \\
Naïve          &            16.342 &             11.61 &           15.256 &            9.161 &            3.045 &            43.003 \\
sNaïve         &            16.342 &            12.521 &           15.988 &            9.161 &            3.045 &            13.912 \\
RF-t-s         &             17.31 &            12.321 &           16.008 &            9.009 &            3.321 &            15.797 \\
RNN            &            17.307 &            12.033 &           16.643 &           11.266 &            3.996 &            13.326 \\
XGB-s          &            17.729 &            12.633 &           17.158 &           10.201 &            3.512 &  \bfseries 10.912 \\
XGB            &            17.729 &            12.671 &           17.259 &           10.201 &            3.512 &            14.288 \\
LR-s           &                 - &            12.235 &           17.768 &            9.519 &             3.32 &            11.235 \\
LR             &                 - &            12.238 &           17.797 &            9.519 &             3.32 &            16.342 \\
XGB-t-s        &            17.954 &             13.21 &           18.102 &            9.462 &            3.608 &            17.195 \\
MLP            &            19.166 &             14.04 &           17.625 &            10.12 &            3.568 &            13.211 \\
KNN-t-s        &             18.37 &            13.701 &           19.711 &           10.482 &            4.062 &            16.333 \\
KNN-s          &            18.758 &            13.548 &           19.764 &           11.167 &             4.08 &            11.988 \\
KNN            &            18.758 &            13.606 &           19.905 &           11.167 &             4.08 &             14.61 \\
LR-t-s         &                 - &                 - &           18.619 &  \bfseries 8.567 &            3.363 &             16.76 \\
\bottomrule
\end{tabular}
 % load tabular from file
    \begin{tablenotes}
    \setlength{\itemindent}{-2.49997pt} % remove indent from \item below
    \footnotesize
    \item \emph{Notes:} Rows show forecasters described in table \ref{tab:m4-models-extended}. Columns show M4 data sets grouped by sampling frequency. \InsertLRTableNotes
    \end{tablenotes}
    \end{threeparttable}
\end{table}

%%%%%%%%%%%%%%%%%%%%%%%%%%%%%%%%%%%%%%%%%%%%%%%%%%%%%%%%%%%%%%%%%%%%%%%%%%%%%%%%%%%%
\begin{table}[htbp]
    \begin{threeparttable}
    \centering \small
    \caption{Complete results (MASE)}
    \label{tab:mase_extended}
    \sisetup{
      table-align-uncertainty=true,
      separate-uncertainty=true,
    }
    \begin{tabular}{lllllll}
\toprule
{} &           Yearly &        Quarterly &          Monthly &          Weekly &           Daily &          Hourly \\
\midrule
Theta-bc       &  \bfseries 2.951 &            1.186 &            0.964 &           2.582 &           3.252 &           2.465 \\
Theta          &            3.279 &            1.218 &            0.968 &           2.637 &           3.261 &           2.457 \\
RF-Theta-bc-t  &            3.246 &            1.229 &            0.985 &           2.592 &           3.257 &           2.363 \\
ARIMA          &            3.407 &            1.204 &  \bfseries 0.939 &  \bfseries 2.27 &           3.244 &           0.981 \\
RF-Theta-bc    &            3.282 &            1.253 &            0.984 &            2.87 &  \bfseries 3.18 &           2.365 \\
Com            &            3.371 &  \bfseries 1.168 &            0.976 &           2.386 &           3.213 &           4.383 \\
Damped         &            3.504 &  \bfseries 1.168 &            0.987 &           2.334 &           3.262 &           2.922 \\
XGB-Theta-bc   &            3.392 &            1.338 &            1.025 &           3.127 &           3.309 &           2.387 \\
XGB-Theta-bc-t &            3.476 &            1.332 &            1.032 &           2.713 &           3.446 &           2.398 \\
KNN-Theta-bc   &            3.394 &            1.366 &            1.069 &            3.01 &           3.365 &            2.52 \\
KNN-Theta-bc-t &            3.396 &            1.368 &            1.083 &           2.842 &             3.4 &           2.529 \\
RF-s           &            3.639 &            1.327 &            1.016 &           2.809 &           3.523 &  \bfseries 0.93 \\
RF             &             3.64 &            1.341 &            1.037 &           2.823 &           3.521 &           1.032 \\
Holt           &            3.748 &            1.198 &            1.038 &           2.381 &            3.23 &           8.988 \\
RF-t-s         &            3.779 &            1.399 &            1.068 &           2.639 &           3.553 &           1.212 \\
SES            &             3.98 &             1.34 &             1.02 &           2.684 &           3.281 &           2.385 \\
RNN            &            3.733 &            1.329 &            1.116 &           3.139 &           3.962 &           2.408 \\
Naïve2         &            3.974 &            1.371 &            1.063 &           2.777 &           3.278 &           2.395 \\
XGB-s          &            3.814 &             1.42 &            1.113 &           3.091 &           3.754 &           0.954 \\
XGB            &            3.814 &            1.431 &            1.129 &           3.091 &           3.754 &           1.063 \\
XGB-t-s        &            3.883 &            1.493 &            1.154 &           2.857 &           3.847 &           1.335 \\
Naïve          &            3.974 &            1.477 &            1.205 &           2.777 &           3.278 &          11.608 \\
sNaïve         &            3.974 &            1.602 &             1.26 &           2.777 &           3.278 &           1.193 \\
MLP            &            4.221 &            1.615 &            1.129 &           2.694 &           3.763 &            2.35 \\
KNN-s          &            4.072 &            1.524 &            1.217 &           3.378 &            4.38 &           1.044 \\
KNN-t-s        &            3.977 &            1.543 &            1.261 &           3.192 &           4.355 &           1.471 \\
KNN            &            4.072 &            1.534 &            1.239 &           3.378 &            4.38 &           1.112 \\
LR-s           &                - &            1.319 &            3.284 &           2.467 &           3.442 &           0.934 \\
LR             &                - &            1.327 &            3.301 &           2.467 &           3.442 &           1.001 \\
LR-t-s         &                - &                - &            2.618 &           2.338 &           3.518 &           1.027 \\
\bottomrule
\end{tabular}
 % load tabular from file
    \begin{tablenotes}
    \setlength{\itemindent}{-2.49997pt} % remove indent from \item below
    \footnotesize
    \item \emph{Notes:} Rows show forecasters described in table \ref{tab:m4-models-extended}. Columns show M4 data sets grouped by sampling frequency. \InsertLRTableNotes
    \end{tablenotes}
    \end{threeparttable}
\end{table}

%%%%%%%%%%%%%%%%%%%%%%%%%%%%%%%%%%%%%%%%%%%%%%%%%%%%%%%%%%%%%%%%%%%%%%%%%%%%%%%%%%%%
\begin{table}[htbp]
    \begin{threeparttable}
    \centering \small
    \caption{OWA}
    \label{tab:owa_extended}
    \sisetup{
      table-align-uncertainty=true,
      separate-uncertainty=true,
    }
    \begin{tabular}{lllllll}
\toprule
{} &           Yearly &        Quarterly &          Monthly &           Weekly &            Daily &           Hourly \\
\midrule
Theta-bc       &  \bfseries 0.776 &            0.893 &  \bfseries 0.904 &            0.964 &            0.996 &            1.009 \\
Theta          &            0.852 &            0.912 &            0.906 &            0.972 &            0.999 &            1.006 \\
Com            &            0.886 &  \bfseries 0.885 &             0.93 &            0.911 &  \bfseries 0.982 &            1.506 \\
RF-Theta-bc-t  &            0.854 &            0.938 &            0.933 &            0.959 &            0.999 &            0.987 \\
ARIMA          &              0.9 &            0.934 &            0.919 &  \bfseries 0.885 &            1.008 &            0.577 \\
Damped         &            0.913 &            0.886 &            0.931 &            0.901 &            1.006 &             1.13 \\
RF-Theta-bc    &            0.864 &            0.957 &            0.932 &            1.052 &            0.988 &            0.985 \\
XGB-Theta-bc   &            0.898 &            1.017 &            0.971 &            1.153 &            1.023 &            0.993 \\
SES            &            1.002 &             0.97 &            0.952 &            0.975 &              1.0 &             0.99 \\
XGB-Theta-bc-t &            0.906 &            1.009 &            0.976 &            1.006 &             1.04 &            0.998 \\
RF-s           &            0.967 &            1.014 &            0.994 &            1.015 &            1.078 &  \bfseries 0.493 \\
Holt           &            0.992 &            0.924 &            1.021 &            0.941 &            0.997 &            2.637 \\
KNN-Theta-bc   &            0.901 &            1.045 &            1.016 &            1.149 &            1.037 &            1.052 \\
Naïve2         &              1.0 &              1.0 &              1.0 &              1.0 &              1.0 &              1.0 \\
KNN-Theta-bc-t &            0.902 &            1.045 &            1.025 &            1.067 &            1.042 &            1.055 \\
RF             &            0.967 &            1.025 &            1.008 &            1.016 &            1.077 &            0.587 \\
RF-t-s         &            1.005 &             1.07 &            1.057 &            0.967 &            1.087 &            0.683 \\
RNN            &            0.999 &            1.031 &            1.102 &             1.18 &             1.26 &            0.865 \\
Naïve          &              1.0 &            1.066 &            1.095 &              1.0 &              1.0 &            3.593 \\
XGB-s          &            1.022 &            1.091 &            1.118 &            1.113 &            1.149 &            0.496 \\
XGB            &            1.022 &            1.097 &            1.129 &            1.113 &            1.149 &            0.611 \\
sNaïve         &              1.0 &            1.153 &            1.146 &              1.0 &              1.0 &            0.628 \\
XGB-t-s        &            1.038 &            1.144 &             1.17 &            1.031 &            1.179 &            0.746 \\
MLP            &            1.117 &            1.226 &            1.142 &            1.037 &             1.16 &             0.85 \\
KNN-s          &            1.086 &            1.171 &            1.257 &            1.218 &            1.338 &            0.544 \\
KNN-t-s        &            1.062 &            1.185 &            1.276 &            1.147 &            1.331 &            0.751 \\
KNN            &            1.086 &            1.177 &            1.272 &            1.218 &            1.338 &             0.63 \\
LR-s           &                - &            1.037 &             2.16 &            0.964 &             1.07 &            0.501 \\
LR             &                - &            1.039 &            2.169 &            0.964 &             1.07 &            0.654 \\
LR-t-s         &                - &                - &            1.876 &            0.889 &            1.089 &             0.67 \\
\bottomrule
\end{tabular}
 % load tabular from file
    \begin{tablenotes}
    \setlength{\itemindent}{-2.49997pt} % remove indent from \item below
    \footnotesize
    \item \emph{Notes:} Rows show forecasters described in table \ref{tab:m4-models-extended}. Columns show M4 data sets grouped by sampling frequency. \InsertLRTableNotes
    \end{tablenotes}
    \end{threeparttable}
\end{table}

%%%%%%%%%%%%%%%%%%%%%%%%%%%%%%%%%%%%%%%%%%%%%%%%%%%%%%%%%%%%%%%%%%%%%%%%%%%%%%%%%%%%
\begin{table}[htbp]
    \begin{threeparttable}
    \centering \small
    \caption{Summary results of new machine learning models}
    \label{tab:overall_extended}
    \sisetup{
      table-align-uncertainty=true,
      separate-uncertainty=true,
    }
    \begin{tabular}{lllll}
\toprule
{} &             sMAPE &             MASE &              OWA & Running time (min) \\
\midrule
Theta-bc       &  \bfseries 11.952 &  \bfseries 1.583 &  \bfseries 0.876 &                8.1 \\
Theta          &            12.264 &            1.669 &              0.9 &               6.27 \\
Com            &            12.668 &            1.687 &            0.914 &              69.47 \\
RF-Theta-bc-t  &            12.673 &            1.671 &            0.919 &           14539.77 \\
ARIMA          &            12.992 &            1.673 &             0.92 &           14992.88 \\
Damped         &            12.692 &            1.718 &             0.92 &              53.45 \\
RF-Theta-bc    &            12.763 &            1.682 &            0.925 &            1189.37 \\
XGB-Theta-bc   &            13.357 &            1.754 &            0.968 &             122.94 \\
SES            &             13.09 &            1.885 &             0.97 &                5.9 \\
XGB-Theta-bc-t &            13.337 &             1.78 &            0.971 &            1826.44 \\
RF-s           &            14.002 &            1.806 &            0.994 &            1168.56 \\
Holt           &             14.16 &             1.83 &            0.997 &              11.72 \\
KNN-Theta-bc   &            13.818 &            1.785 &            0.998 &              64.36 \\
Naïve2         &            13.564 &            1.912 &              1.0 &               3.66 \\
KNN-Theta-bc-t &            13.848 &            1.794 &            1.003 &             656.04 \\
RF             &            14.095 &             1.82 &            1.004 &            1163.64 \\
RF-t-s         &             14.86 &            1.882 &            1.047 &           14165.87 \\
RNN            &            15.122 &            1.902 &            1.067 &           38941.68 \\
Naïve          &            14.208 &            2.044 &            1.072 &               1.04 \\
XGB-s          &            15.576 &            1.926 &            1.088 &             116.95 \\
XGB            &            15.647 &            1.937 &            1.096 &             115.76 \\
sNaïve         &            14.657 &            2.057 &            1.105 &     \bfseries 1.03 \\
XGB-t-s        &            16.246 &            1.983 &            1.131 &            1718.15 \\
MLP            &             16.48 &            2.079 &            1.156 &             157.88 \\
KNN-s          &            17.315 &            2.088 &            1.197 &              58.71 \\
KNN-t-s        &            17.252 &            2.092 &            1.205 &             634.12 \\
KNN            &            17.407 &            2.101 &            1.207 &              56.99 \\
LR-s           &                 - &                - &                - &              55.36 \\
LR             &                 - &                - &                - &              53.37 \\
LR-t-s         &                 - &                - &                - &             590.87 \\
\bottomrule
\end{tabular}
 % load tabular from file
    \begin{tablenotes}
    \setlength{\itemindent}{-2.49997pt} % remove indent from \item below
    \footnotesize
    \item \emph{Notes:} Rows show forecasters described in table \ref{tab:m4-models-extended}. Columns show aggregate values for sMAPE, MASE and OWA metrics, as well as the total running time in minutes scaled to the number of CPUs used in the original M4 study, as described in section \ref{sec:experiments}. \InsertLRTableNotes
    \end{tablenotes}
    \end{threeparttable}
\end{table}

%%%%%%%%%%%%%%%%%%%%%%%%%%%%%%%%%%%%%%%%%%%%%%%%%%%%%%%%%%%%%%%%%%%%%%%%%%%%%%%%%%%%
%%%%%%%%%%%%%%%%%%%%%%%%%%%%%%%%%%%%%%%%%%%%%%%%%%%%%%%%%%%%%%%%%%%%%%%%%%%%%%%%%%%%
\begin{table}[htbp]
    \begin{threeparttable}
    \centering \small
    \caption{Post-hoc Wilcoxon signed rank tests results for pairwise comparisons on the hourly data set}
    \label{tab:wilcoxon_holm_q1}
    \begin{tabular}{lllr}
\toprule
{} &  forecaster A & forecaster B &  p-value \\
\midrule
0 &          LR-s &         RF-s &    0.657 \\
4 &          LR-s &        XGB-s &    0.243 \\
3 &  M4 runner-up &    M4 winner &    0.317 \\
2 &  M4 runner-up &         RF-s &    0.468 \\
1 &          RF-s &        XGB-s &    0.010 \\
\bottomrule
\end{tabular}
 % load tabular from file
    \begin{tablenotes}
    \setlength{\itemindent}{-2.49997pt} % remove indent from \item below
    \footnotesize
    \item \emph{Notes:} This table corresponds to the results presented in figure \ref{fig:cd_q1}. The results are based on sMAPE and the hourly data set. We only show pairs of forecasters which are not significantly different. Forecasters are described in table \ref{tab:m4-models-extended}. We use Wilcoxon signed rank tests to compute p-values. Significance at the $5$\% is established using Holm's procedure for correcting for multiple testing, as discussed in \ref{sec:evaluation}. 
    \end{tablenotes}
    \end{threeparttable}
\end{table}

%%%%%%%%%%%%%%%%%%%%%%%%%%%%%%%%%%%%%%%%%%%%%%%%%%%%%%%%%%%%%%%%%%%%%%%%%%%%%%%%%%%%
%%%%%%%%%%%%%%%%%%%%%%%%%%%%%%%%%%%%%%%%%%%%%%%%%%%%%%%%%%%%%%%%%%%%%%%%%%%%%%%%%%%%
\begin{table}[htbp]
    \begin{threeparttable}
    \centering
    \small
    \caption{Post-hoc Wilcoxon signed rank tests results for pairwise comparisons on the hourly data set}
    \label{tab:wilcoxon_holm_q2}
    \begin{tabular}{lllr}
\toprule
{} &   forecaster A &    forecaster B &  p-value \\
\midrule
2 &    RF-Theta-bc &   RF-Theta-bc-t &    0.096 \\
1 &  RF-Theta-bc-t &    XGB-Theta-bc &    0.092 \\
3 &       Theta-bc &  XGB-Theta-bc-t &    0.527 \\
0 &   XGB-Theta-bc &  XGB-Theta-bc-t &    0.028 \\
\bottomrule
\end{tabular}
 % load tabular from file
    \begin{tablenotes}
    \setlength{\itemindent}{-2.49997pt} % remove indent from \item below
    \footnotesize
    \item \emph{Notes:} This table corresponds to the results presented in figure \ref{fig:cd_q2}. The results are based on sMAPE and the hourly data set. We only show pairs of forecasters which are not significantly different. Forecasters are described in table \ref{tab:m4-models-extended}. We use Wilcoxon signed rank tests to compute p-values. Significance at the $5$\% is established using Holm's procedure for correcting for multiple testing, as discussed in \ref{sec:evaluation}. 
    \end{tablenotes}
    \end{threeparttable}
\end{table}

%%%%%%%%%%%%%%%%%%%%%%%%%%%%%%%%%%%%%%%%%%%%%%%%%%%%%%%%%%%%%%%%%%%%%%%%%%%%%%%%%%%%
%%%%%%%%%%%%%%%%%%%%%%%%%%%%%%%%%%%%%%%%%%%%%%%%%%%%%%%%%%%%%%%%%%%%%%%%%%%%%%%%%%%%
\end{document}